\documentclass[runningheads]{llncs}

\usepackage{graphicx}

\usepackage{tikz}
\usepackage{hyperref}
\usepackage{comment}
\usepackage{amsmath,amssymb} 
\usepackage{color}

\usepackage[accsupp]{axessibility}  

\usepackage{algorithm}
\usepackage{algorithmic}
\usepackage{mathtools}
\usepackage{multirow}
\usepackage{hhline}
\usepackage{adjustbox}
\usepackage{xcolor}

\usepackage{amsmath}
\usepackage{amsfonts}

\newcommand\Scal{\mathcal{S}}

\begin{document}
\pagestyle{headings}
\mainmatter


\title{GEMS: Scene Expansion using Generative Models of Graphs}

\makeatletter
\renewcommand*{\@fnsymbol}[1]{\ensuremath{\ifcase#1\or *\or \dagger\or \ddagger\or
   \mathsection\or \mathparagraph\or \|\or **\or \dagger\dagger
   \or \ddagger\ddagger \else\@ctrerr\fi}}
\makeatother

\titlerunning{GEMS}
%
\author{Rishi Agarwal\inst{1}\thanks{The first four authors contributed equally to this work}\thanks{The first three authors contributed to the work during their Adobe internship.} \and
Tirupati Saketh Chandra\inst{1*\dagger} \and
Vaidehi Patil\inst{1*\dagger} \and \\
Aniruddha Mahapatra\inst{2*} \and
Kuldeep Kulkarni\inst{2} \and
Vishwa Vinay\inst{2}}
\authorrunning{Agarwal et al.}
%
\institute{Indian Institute of Technology, Bombay
\email{\{rishiapril2000,tsakethchandra,vaidehipatil16\}@gmail.com} \and
Adobe Research \\
\email{\{anmahapa,kulkulka,vinay\}@adobe.com}}
\maketitle

\begin{abstract}
Applications based on image retrieval require editing and associating in intermediate spaces that are representative of the high-level concepts like objects and their relationships rather than dense, pixel-level representations like RGB images or semantic-label maps. We focus on one such representation, scene graphs, and propose a novel scene expansion task where we enrich an input seed graph by adding new nodes (objects) and the corresponding relationships. To this end, we formulate scene graph expansion as a sequential prediction task involving multiple steps of first predicting a new node and then predicting the set of relationships between the newly predicted node and previous nodes in the graph. We propose a sequencing strategy for observed graphs that retains the clustering patterns amongst nodes. In addition, we leverage external knowledge to train our graph generation model,  enabling greater generalization of node predictions. Due to the inefficiency of existing maximum mean discrepancy (MMD) based metrics for graph generation problems in evaluating predicted relationships between nodes (objects), we design novel metrics that comprehensively evaluate different aspects of predicted relations.  We conduct extensive experiments on Visual Genome and VRD datasets to evaluate the expanded scene graphs using the standard MMD based metrics and our proposed metrics. We observe that the graphs generated by our method, GEMS, better represent the real distribution of the scene graphs than the baseline methods like GraphRNN.

\end{abstract}

\section{Introduction}
Creative photographers are gifted with the ability to imagine a set of concepts - objects and inter-object relationships - to capture in a photograph. However, they spend prohibitively expensive amount of time arriving at the kinds of scenes that contain this seed set of concepts they desire to be present in the photograph and are often tasked with sifting through a large number of photos to zero-in the scene they want to capture. Hence, it is desirable to empower them with recommendations of a wide variety of diverse and rich scenes that contain these seed concepts. We wish to devise algorithms that can be leveraged to provide the user effective recommendations of scenes that subsume the seed concepts while ensuring they are diverse and richer than the seed concepts can represent. 

To this end, we express the seed concepts in the form of a scene graph ~\cite{xu2020survey,johnson2015image} and cast the task of producing richer and diverse scenes as the generating of plausible novel scene graphs that subsume the seed graph. Specifically, we propose a novel \textit{scene expansion} problem - given a seed scene graph, can we enhance it by the addition of new objects so that the new graph corresponds to an enriched scene while satisfying the following requirements: (a) the proposed additions respect object co-occurrence patterns observed in the training set; (b) the enhanced scene graph should be novel with respect to the existing collection of graphs; and (c) it should be possible to generate multiple different graphs for the same seed.

The space of generative models for unconditional generation of molecular graphs have received attention \cite{grover2019graphite,simonovsky2018graphvae,JTVAE,samanta2018nevae,molgan} recently.
Specifically, the auto-regressive models \cite{graphgen,graphrnn} that have been shown to work well for molecular graph generation can potentially be repurposed for our task of scene graph expansion. However, the complexity of the graphs considered in these works tend to be several orders of magnitude smaller than scene graphs, in terms of the number of distinct types of nodes and relationship-edges. Moreover, these methods implicitly require the graphs to be connected, which is not necessarily a characteristic of the scene graphs that we deal with. 
In addition, scene graphs tend to be more diverse compared to molecular graphs. For these reasons, the above mentioned auto-regressive models that are proposed for graph generation are unsuitable for the scene graph expansion problem that we tackle. 

Motivated by this, we design a novel auto-regressive graph expansion model, \textbf{GEMS} - Graph Expansion Model for Scenes, drawing inspiration from \cite{graphrnn} that can generate graphs of various lengths unlike \cite{fan2019labeled,simonovsky2018graphvae,Cao2018MolGANAI}. We first convert the scene graphs into sequences where each node in the sequence is separated by relationships with previous nodes in the sequence using our proposed sequencing method that tries to ensure that groups of objects connected in the scene graph occur close by in the sequence ensuring model learns an approximate notion of motifs \cite{zellers2018neural}. We predict the nodes and then edges using using Gated Recurrent Units (GRUs), where node generation precedes edge generation. Due to the imbalance in edge-types in scene graphs we use class-rebalancing loss that drastically improves degenerate predictions of edge-labels. Further, we incorporate external knowledge derived from language domain for better generalization of node predictions to encourage generation of a diverse set of related node predictions instead of the same node predicted multiple times in an expanded graph. Since the problem of scene graph expansion is novel, there are no defined metrics to evaluate the quality of expanded graphs. Hence, we use metrics defined in \cite{graphgen} along with our proposed metrics that are specifically tailored for scene graphs to capture properties not done by previous metrics.

\paragraph{Contributions:}
We summarize our key contributions.
\begin{itemize}
    \item We propose a novel scene expansion task that deals with enhancing a given seed graph by the addition of new objects and relationships so that the enhanced graph corresponds to an enriched scene.
    \item We design an auto-regressive model, GEMS, for conditional generation of scene graphs that generates nodes and edges hierarchically in a dependent fashion.
    \item We propose a novel Cluster-Aware graph sequencing method (Cluster-Aware BFS) aimed at  capturing object co-occurrences and learning common object collections.
    \item We incorporate external knowledge in the form of an additional loss component for better generalization of node predictions.
    \item To circumvent the drawbacks of the traditional evaluation metrics, we propose additional metrics to evaluate the generated scene graphs to capture the \textit{coherence} of predicted edges and nodes.
\end{itemize}
Through extensive experiments on Visual Genome \cite{visualgenome} and VRD \cite{vrd-and-language-priors} dataset, we show that our model outperforms the GraphRNN based baseline models comprehensively in most of the metrics and is somewhat competitive with \cite{garg2021unconditional} that introduces complementary ideas to ours.

\section{Related Work}
The relevant literature is introduced from the following two aspects: ($1$)~Scene Graph Extraction ($2$)~Generative Models of Graphs.
\subsubsection{Scene Graph Extraction}
The standard task known as scene graph \textit{generation} involves constructing a graph with nodes as objects and their attributes with edges being relationships between them. A variant of this, that refers to producing a graph from an image input is referred to ``extraction'' in the rest of the paper. Broadly, scene extraction methods fall into two categories. First one, dubbed in this paper as \textbf{Internal Knowledge}, refers to the works ~\cite{li2017scene,dai2017detecting,zhang2017relationship,klawonn2018generating} that deal with features that are leveraged to produce the graph originate from only the image of interest. At a high level, the scene graph extraction operates by detecting objects and their regions within the image followed by a relationship identification model sub-component that labels the connections between the objects.  
Subsequent works have attempted to address the issue arising from biased nature of the training data caused by the long-tail of less-frequently occurring relationships ~\cite{dornadula2019visual,UnbiasedSGTang2020}. 
The second line of work in scene graph extraction leverages the {\bf external knowledge} in the form of word embeddings  ~\cite{vrd-and-language-priors} for the object and relationship class names as a prior from the language domain. Scene graph extraction methods that combine the internal knowledge within the image and external knowledge have shown increased levels of accuracy~\cite{ext-knowledge-2,ext-knowledge-3}. This reference information, typically in the form of a knowledge graph, aids in scene understanding~\cite{BridgingKGZareian2020}. Similarly, the notion of a common-sense knowledge~\cite{ilievski2021cskg} can potentially be used to aid the scene graph extraction process.

Our work differs from these two lines of works in that we do not have access to the input image to extract the visual features, i.e the internal knowledge and only have an input seed graph. Hence, similar to the second line of work, we leverage the external knowledge both an input level and as well as a regularizer for the additional nodes and edges in the output graph.
We invoke state-of-the-art generative models of graphs, described next, to expand the given seed. 

\subsubsection{Generative Models of Graphs}
Graphs are a powerful and natural representation of the data in many application settings. And, as with many other domains, generative models trained over a set of observed graphs have received much recent attention~\cite{guo2020systematic}. Most existing work considers molecular graphs, where sampling from a trained model allows the generation of novel molecules, the core objective of drug design. Variational Auto-Encoders (VAEs) are a popular method within this class of models~\cite{grover2019graphite,simonovsky2018graphvae,JTVAE,samanta2018nevae}, and so are Generative Adversarial Networks (GANs)~\cite{molgan}.
Papers operating on molecular graphs leverage knowledge about the chemistry of the corresponding molecules. 
In the current paper, we consider scene graphs derived from images, where sparsity~\cite{pmlr-v119-dai20b} needs to be specifically addressed since most object-object pairs do not have a relationship between them. In addition, scene graphs tend to be diverse, a characteristic they share with graphs from few other domains~\cite{liao2019efficient,graphrnn}. In addition, the \textit{external knowledge} mentioned in the earlier section is more latent and difficult to provide as guidance during model training. 

The closest to our work is ScenegraphGen ~\cite{garg2021unconditional} that introduces several complementary ideas. We differentiate our work on two main aspects: ($1$) We define a custom sequencing function that tries to ensure that groups of objects connected in the scene graph occur close by in the sequence. ($2$) Similar to recent work on scene graph extraction from images, we show how the use of external information about object-object similarities can help guide the model towards scene graphs that are more coherent. ($3$) Our exhaustive experimental evaluation also proposes some novel metrics for the scene graph expansion task. 

\section{Problem Description and Model}
\label{sec:modelTrainingAndInference}

We are given a collection of observed scene graphs $\mathbb{G} = \{G\}$ where each $G$ corresponds to an image and is represented by $G=(V,E)$ - a set of vertices $V  \subseteq \mathbb{V}$ and directed, labelled edges $E \subseteq \{(u,e,v) | u,v \in \mathbb{V}, u\ne v, e \in \mathbb{E}\}$ that connect pairs of objects in $V$. Here, $\mathbb{V}$ is the set of distinct objects found in the collection of scene graphs and $\mathbb{E}$ is the set of unique relationships. Our objective is to take a graph $G_s \notin \mathbb{G}$ and expand it into $\hat{G}_s$ such that $G_s$ is a subgraph of $\hat{G}_s$. 
Drawing inspiration from \cite{li2018learning,graphrnn} we convert the graph into a sequence and transform this problem of graph expansion to sequential prediction. That is, under a node ordering $\pi \in \Pi$, a graph $G$ is flattened into a sequence $\Scal(G) = \{(v_i, \mathcal{E}_i)\}_{i=1}^{n}$, where $v_i \in \mathbb{V}$ indexes the $i^{th}$ node in the sequence induced by $\pi$. And $\mathcal{E}_i = \{\mathcal{E}_{i,j}\}_{j<i}$ is a list containing edge information for node $i$ with every node $j$ before it in the sequence. Since scene graphs are directed, we take each $\mathcal{E}_{i,j}=(e_{i,j}, e_{j,i})$ to be a pair of relationships - one in each direction - with $e_{i,j}$ denoting the relationship from $v_i$ to $v_j$.

\subsubsection{Cluster-Aware BFS}
\label{motif-aware-bfs}

Critical to being able to train graph generation models is the role of the process that converts a graph $G$ into a sequence $\Scal(G)$. GraphRNN~\cite{graphrnn} uses a breadth-first strategy (BFS) while GraphGen~\cite{graphgen} uses a depth-first traversal. Both these options require the input graph to be fully connected. The input scene graphs in our context often contain disconnected components, these correspond to natural scenes where part of the image may not have relationships with objects in other parts. Additionally, as we might intuitively expect, some sets of objects co-occur frequently across the dataset. In an attempt to encourage the model to better handle clusters of objects that occur together, we devise a method that ensures that objects in the same cluster are close by in the sequence. 
For any given scene graph, we first identify its maximal connected subgraphs. We obtain the BFS sequence for each subgraph with a randomly chosen node ordering. The sequence for the scene graph is obtained by concatenating subgraph sequences in random order. Alternatives to BFS might be possible, our objective is to help the model learn co-occurrence amongst sets-of-objects. Randomizing across subgraphs before concatenation is aimed at introducing robustness with respect to the input seed graph. 

As previously defined, $\Scal(G)$ can be thought of as a matrix where row $i$ holds information about $v_i$ and its relationships with the previously seen objects. We use the shorthand $\Scal_{i}$ for all information about the $i^{th}$ node in the sequence for graph $G$, and $\Scal_{<i}$ for all nodes and edges occurring before it. A likelihood can now be defined for the sequence:
\begin{equation}
\label{eq:graphSeqLikelihood}
\begin{split}
    P(\Scal(G)) = \prod_{i=1}^{n_G} P(v_i | \Scal_{< i}) \times P(\mathcal{E}_i | \Scal_{<i}, v_i) \\
\end{split}
\end{equation}
\begin{equation}
    P(\mathcal{E}_i | \Scal_{<i}, v_i) = \prod_{j<i} P(\mathcal{E}_{i,j} | \Scal_{<i}, \mathcal{E}_{i,<j}, v_i, v_j)
\end{equation}

\subsubsection{Hierarchical Node and Edge Prediction}
\label{node-and-edge-pred}
The expansion of graph sequence occurs in steps, in each step we first predict a new node $\hat{v}_i$ and then a sequence relationships between node $\hat{v}_i$ and previous nodes in the sequence. In the current paper, both $P(v_i | \Scal_{< i})$ and $P(\mathcal{E}_{i,j} | \Scal_{<i}, \mathcal{E}_{i,<j}, v_i, v_j)$ are modeled separately by recurrent neural networks, given by $f_{node}$ and $f_{edge}$ respectively, with the corresponding parameters shared across different steps. Prediction of $i^{th}$ node $\hat{v}_i$ is defined as,
\begin{equation}
    \hat{v}_i \sim f_{node} (\Scal_{i-1}, h_{node}(\Scal_{<i}))
\end{equation}
That is, the prediction of the $i^{th}$ node in the sequence depends on $\Scal_{i-1}$ and the hidden state of $f_{node}$ from previous step. Correspondingly, the prediction of the new  edge pair ($\hat{e}_{i,j}, \hat{e}_{j,i}$) is given by,
\begin{equation}
    \hat{e}_{i,j} \sim f_{edge}(v_i, v_j, h_{edge}(\Scal_{<i}, \mathcal{E}_{i,<j}))
\end{equation}
\begin{equation}
    \hat{e}_{j,i} \sim f_{edge}(v_j, v_i, h_{edge}(\Scal_{<i}, \mathcal{E}_{i,<j}, \hat{e}_{i,j}))
\end{equation} 
Note that in our formulation we first predict $e_{i,j}$ and then  $e_{j,i}$ to get $\mathcal{\hat{E}}_{i,j}$. For edge prediction, we explicitly provide $v_i$ and $v_j$ as inputs into $f_{edge}$ because the existence of an edge between two nodes, as well as its label, is more dependent on the local context (the nodes) than the rest of the graph.
The objects and relationships are sampled from multinomial distributions, making our model a Dependent Multinomial Sequence Model (rather than Bernoulli sequences as in~\cite{graphrnn}).
Our training objective is a combination of the losses computed on node and edge prediction:
\begin{equation}
\begin{aligned}
    \mathcal{L}(G) = \sum_{i \in V} l_{node}(p_{v_{i}}, p_{\hat{v}_i}) + \sum_{j < i; (v_i, e, v_j) \in E} l_{edge}(p_{e_{i,j}}, p_{\hat{e}_{i,j}}) \\
    + \sum_{j < i; (v_j, e, v_i) \in E} l_{edge}(p_{e_{j,i}}, p_{\hat{e}_{j,i}})
\end{aligned}
\end{equation}  
We define the node prediction loss as:
\begin{equation}
\label{eq:nodeLoss}
\begin{aligned}
    l_{node}(p_{v_{i}}, p_{\hat{v}_i}) &= H (p_{v_{i}}, p_{\hat{v}_i})
\end{aligned}
\end{equation}  
Where, $H$ is the cross-entropy loss between a 1-hot encoding for the node label of $v_i$ ($p_{v_{i}}$) and the corresponding predicted probability $p_{\hat{v}_i}$ for $\hat{v}_i$ . 
Similarly, the edge prediction is defined as,
\begin{equation}
\label{eq:edgeLoss}
    l_{edge}(p_{e_{i,j}}, p_{\hat{e}_{i,j}}) = \frac{1-\beta}{1-\beta^{N_e}} H (p_{e_{i,j}}, p_{\hat{e}_{i,j}})
\end{equation}
Where, $p_{e_{i,j}}$ is the 1-hot encoding of the ground-truth edge type, $p_{\hat{e}_{i,j}}$ is the is the probability distribution of predicted edge type between $v_i$ and $v_j$ and $N_e$ is the number of instances of this edge across the dataset. The edge loss is a class balanced loss \cite{cui2019class} designed to reduce the effect of a highly skewed distribution across relationship classes~\cite{UnbiasedSGTang2020} to produce a model which is less prone to predicting degenerate edges.

\subsubsection{External Knowledge}
\label{external-knowledge}
Cross-entropy is a very strict loss, in the sense that near misses (predicting a node that is similar but not the ground-truth node) are not considered different from obvious errors. To encourage this generalization for predicted node labels we add an additional loss term $H(p_{\hat{v}_i}, q_i)$, where, 
\begin{equation}
\label{eq:external}
\begin{aligned}
    q_i &= \min_{q} KL(q,p_{\hat{v}_i}) - E_{v \sim q}[f(v, v_i)]
\end{aligned}
\end{equation} 
$f(v_i, \hat{v}_i)$ is the similarity between $v_i$ and $\hat{v}_i$ as obtained from external knowledge, and KL is Kullback-Leibler Divergence.
$q_i$ enables the model to predict a $\hat{v}_i$ that may be different from the ground-truth $v_i$ but similar to a proxy $q_i$~\cite{ext-knowledge-1,ext-knowledge-2}.
The node prediction loss thus becomes:
\begin{equation}
\label{eq:nodeLoss_final}
\begin{aligned}
    l_{node}(p_{v_{i}}, p_{\hat{v}_i}) &= (1-\alpha) H (p_{v_{i}}, p_{\hat{v}_i}) + \alpha H (p_{v_{i}}, q_i)
\end{aligned}
\end{equation}  
where $\alpha$ is a hyperparameter.

\subsubsection{Inference} We convert the input seed graph $G_s$ into a sequence $\Scal(G_s)$. Using GEMS, we extend the sequence by progressively adding nodes and edges. To add a new node, we compute the distribution over node labels using the network $f_{node}$ and sample from this multinomial distribution. To add a relationship between nodes $v_i$ and $v_j$, we pick the most probable edge label between the two nodes as predicted by $f_{edge}$. In this way, the seed graph $G_s$ is sequentially expanded to provide $\hat{G}_s$, corresponding to an enhanced scene. 


\section{Experiments}
In this section, we provide empirical validation for the method described earlier. We begin by outlining the dataset and experiment design used.

\begin{table*}[t]
\centering
\begin{adjustbox}{ width=\textwidth}
\renewcommand{\arraystretch}{1.3}
\begin{tabular}{|c|c|c|c|c|c|c|c|c|c|}
\hline 
 & & \multicolumn{4}{|l|}{\hspace{2cm}\textbf{Visual Genome}} & \multicolumn{4}{|l|}{\hspace{2.8cm}\textbf{VRD}} \\
\hhline{~~|--------}
 & & \textbf{GraphRNN} & \textbf{GraphRNN*} & \textbf{SceneGraphGen} & \textbf{GEMS} & \textbf{GraphRNN} & \textbf{GraphRNN**} & \textbf{SceneGraphGen} & \textbf{GEMS} \\
\hline \hline
\multirow{6}{*}{MMD} & Degree (x$10^2$) $\downarrow$ & 47.47 & 16.44 & \textcolor{blue}{\textbf{6.97}} & \textcolor{red}{\textbf{2.11}} & 10.94 & \textcolor{blue}{\textbf{7.94}} & \textcolor{red}{\textbf{4.44}} & 9.91 \\
& Clustering (x$10^2$) $\downarrow$ & 18.63 & 4.05 & \textcolor{red}{\textbf{0.26}} & \textcolor{blue}{\textbf{0.86}} & 16.08 & \textcolor{red}{\textbf{1.21}} & \textcolor{blue}{\textbf{2.58}} & 3.96 \\
& NSPDK* (x$10^3$) $\downarrow$ & 22.60 & 5.10 & \textcolor{red}{\textbf{0.73}} & \textcolor{blue}{\textbf{1.21}} & 6.62 & 6.57 & \textcolor{red}{\textbf{2.85}} & \textcolor{blue}{\textbf{4.09}} \\
& Node Label (x$10^4$) $\downarrow$ & 5.44 & \textcolor{blue}{\textbf{5.26}} & 5.92 & \textcolor{red}{\textbf{5.19}} & 30.70 & 27.01 & \textcolor{red}{\textbf{24.82}} & \textcolor{blue}{\textbf{25.11}} \\
& Edge Label (x$10^2$) $\downarrow$ & 22.38 & 6.19 & \textcolor{red}{\textbf{0.83}} & \textcolor{blue}{\textbf{1.13}} & 1.06 & \textcolor{red}{\textbf{0.39}} & \textcolor{blue}{\textbf{0.60}} & 1.97 \\
\hline
\hline
\multirow{3}{*}{Node Metrics} & Count Reference & \multicolumn{4}{|c|}{11.09} & \multicolumn{4}{|c|}{7.01} \\
\hhline{~|-|----|----}
& Count Predicted & 29.53 & 13.84 & 7.89 & 10.17 & 7.24 & 7.59 & 6.41 & 7.81 \\
& $(Obj)_{K}$ (x$10^2$) $\uparrow$ & 83.7 & 86.9 & \textcolor{red}{\textbf{93.1}} & \textcolor{blue}{\textbf{92.9}} & 87.7 & 85.9 & \textcolor{red}{\textbf{93.1}} & \textcolor{blue}{\textbf{91.1}} \\
\hline
\hline
\multirow{3}{*}{Edge Metrics} & Count Reference & \multicolumn{4}{|c|}{5.01} & \multicolumn{4}{|c|}{7.11} \\
\hhline{~|-|----|----}
& Count Predicted & 57.95 & 11.86 & 5.15 & 7.45 & 10.64 & 6.39 & 7.03 & 8.52 \\
& MEP (x$10^2$) $\uparrow$ & 22.4 & 24.52 & \textcolor{red}{\textbf{53.50}} & \textcolor{blue}{\textbf{35.81}} & 17.94 & 12.40 & \textcolor{red}{\textbf{37.43}} & \textcolor{blue}{\textbf{27.22}} \\
\hline
\hline
\multicolumn{2}{|c|}{Novelty (x$10^2$) $\uparrow$} & 12.26 & 57.59 & \textcolor{red}{\textbf{87.37}} & \textcolor{blue}{\textbf{75.75}} & \textcolor{blue}{\textbf{20.22}} & \textcolor{red}{\textbf{25.39}} & 9.48 & 12.98\\
\hline
\multicolumn{2}{|c|}{Diversity (x$10^2$) $\uparrow$} & 96.70 & \textcolor{red}{\textbf{98.73}} & 79.62 & \textcolor{blue}{\textbf{91.68}} & 90.92 & \textcolor{red}{\textbf{93.73}} & 75.09 & \textcolor{blue}{\textbf{88.80}}\\
\hline
\end{tabular}
\end{adjustbox}
\vspace{0.25cm}
\caption{Comparison of our method (GEMS) with the baseline of \cite{graphrnn} and \cite{scenegraphgen2021} on Visual Genome \cite{visualgenome} and VRD \cite{vrd-and-language-priors} datasets. For all MMD based metrics, lower is better $(\downarrow)$. For the rest of the metrics, larger is better $(\uparrow)$. GraphRNN* and GraphRNN** referes to GraphRNN with $max\_prev\_node=6$ and $7$ for Visual Genome and VRD respectively. \textbf{Note:} \textcolor{red}{\textbf{red}} represents best and \textcolor{blue}{\textbf{blue}} represents second best scores.}
\label{table:baselineResults}
\end{table*}

\vspace{-0.25cm}
\subsubsection{Datasets}
We use two standardized available datasets that have scene graph information. For \textbf{Visual Genome}, we utilize the publicly released preprocessed data from
~\cite{Xu_2017_CVPR}, containing $150$ object classes and $50$ relation classes.
The dataset contains contains human-annotated scene graphs on $108,077$ images. Each image has a scene graph containing on average $11.09$ objects and $5.01$ relationships. We use $70\%$ of the images for training and validation, and the remaining $30\%$ for testing. Similarly, the \textbf{Visual Relationship Dataset(VRD)} dataset contains $100$ object classes and $70$ relation types. The dataset includes $5000$ images, of which we use $80\%$ of the images for training and validation, and the remaining $20\%$ are retained for testing.

\begin{algorithm}
\caption{Extraction of Seed Graph}
\label{seed-graph}
\textbf{Input}: Scene Graph $G$\\
\textbf{Parameter}: k $\leftarrow$ Number of seed graphs\\
\textbf{Output}:
\begin{algorithmic} 
\STATE $S \leftarrow$ set of maximal connected components in $G$
\STATE $seedgraphs$ = []
\FOR{$g \in S$}
\STATE $PR \leftarrow $ empty dictionary
\FOR{$n \in nodes(g)$}
\STATE $PR[n] = PageRank(n)$ in $g$
\ENDFOR
\STATE $subG \leftarrow$ set of all subgraphs of $g$ 
\STATE $pr \leftarrow $ empty list
\FOR{i in range(len($subG$))}
\STATE $pr[i]$ = $\dfrac{1}{\lvert nodes(subG[i]) \rvert}\sum_{n \in nodes(subG[i])}PR[n]$ 
\ENDFOR
\STATE $X \sim Normalize(pr)$
\STATE $seedgraphs[g] \leftarrow k$ samples from  $X$
\ENDFOR
\\ \textbf{return} $seedgraphs$
\end{algorithmic}
\end{algorithm}
\setlength{\textfloatsep}{0.25cm}

\vspace{-0.25cm}
\subsubsection{Implementation Details}
The two RNNs, $f_{node}$ and $f_{edge}$, are implemented as $4$ layers of GRU cells.
We use teacher forcing~\cite{teacher-forcing} during training time, where the ground-truth of observed sequences (nodes \& edges) are used, but model predictions are used during inference. Model fitting utilizes Stochastic Gradient Descent with Adam Optimizer and minibatch of size $32$, with the learning rate set to $0.001$. $\alpha$ and $\beta$ are set to 0.2 and 0.9999 respectively in all experiments.
We use pre-trained GloVe embeddings ~\cite{pennington2014glove} as inputs into the model for node and edge labels. 
For each new predicted node, GEMS predicts edges with $k$-previous nodes denoted by $max\_prev\_node$. We calculate the value of $max\_prev\_node$ empirical by taking that value which covers $99^{th}$ percentile of all graphs in the dataset. Note that this leads to a loss of information (some relationships are ignored), and is an efficiency trade-off. The value of $max\_prev\_node$ for Visual Genome and VRD used are $6$ and $7$ respectively. More details are provided in supplementary. For evaluation, we use the test set to obtain seed graphs and their expanded scenes. We sample subgraphs from a test scene graph to obtain seed graphs, and the scene graph itself plays the role of an expanded scene. The sampling strategy is described in Algorithm~\ref{seed-graph}. We use $k=1$ in Algorithm \ref{seed-graph}.

\begin{figure*}
\begin{center}
\includegraphics[width=\linewidth]{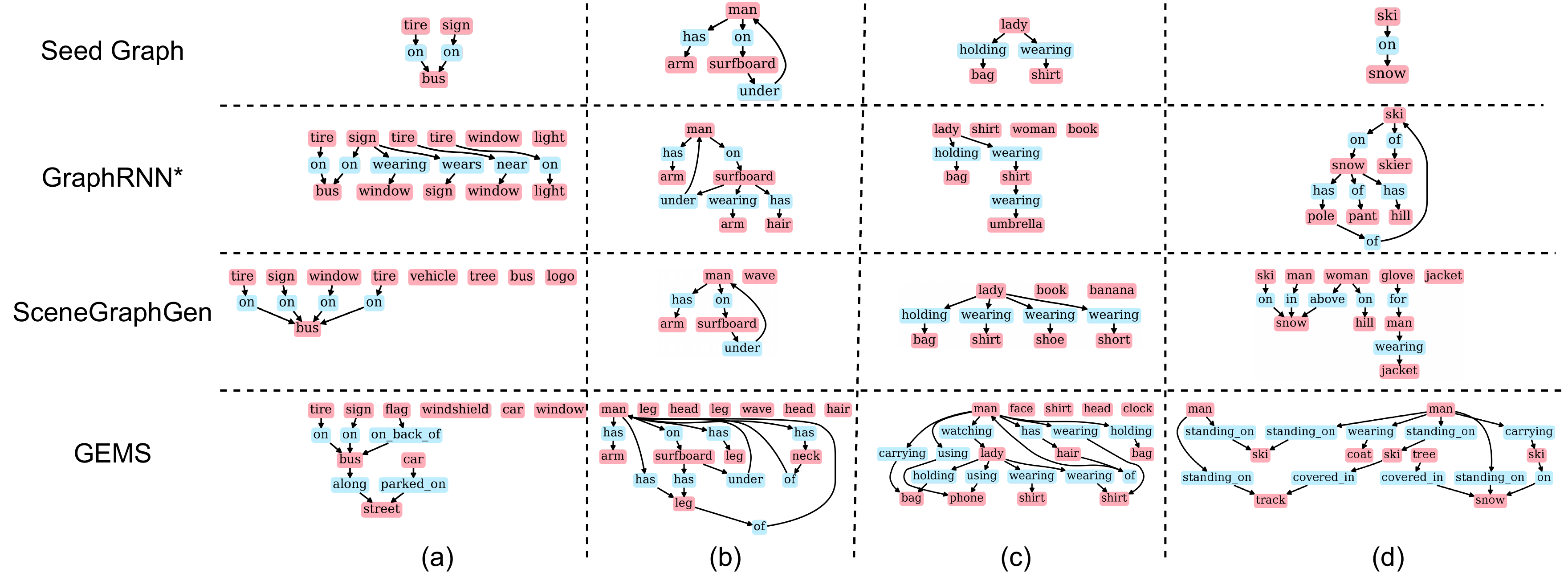}
\end{center}
    \vspace{-0.5cm}
  \caption{Comparison of expanded graphs generated by our model (GEMS) v/s baseline GraphRNN* (GraphRNN with $max\_prev\_node=6$) and SceneGraphGen on Visual Genome seed graphs. Our model generates plausible relationships between objects, while GraphRNN* sometimes predicts irrelevant relationships.}
\label{fig:baselines}
\end{figure*}

\vspace{-0.15cm}

\subsubsection{Baselines}
Since scene graph expansion is a novel task there are no prior baselines for this task. For our purpose, we transform GraphRNN \cite{graphrnn} to work for scene graphs containing bi-directional edge-relations between nodes. Another variant of GraphRNN is used as a baseline, GraphRNN* where $max\_prev\_node$ is set to 6 and 7 respectively for Visual Genome and VRD. Having a smaller value of $max\_prev\_node$ that covers most of the edges in the graph allows the model to focus on more important edges and avoids the prediction of degenerate edges. Additionally, we observe that this helps the model generate nodes and edges much similar in the count to reference training set. We also compare our method against SceneGraphGen \cite{scenegraphgen2021}, which, though mainly focuses on the unconditional generation of scene graphs, can be leveraged for scene graph expansion also. More details of how we transform GraphRNN to work on scene graphs and SceneGraphGen's implementation are provided in the supplementary.

\begin{table*}[t]
\centering
\begin{adjustbox}{ width=0.7\textwidth}
\renewcommand{\arraystretch}{1.3}
\begin{tabular}{|c|c|c|c|c|c|}
\hline 
 & & \textbf{GraphRNN*} & \multicolumn{1}{c|}{\begin{tabular}[c]{@{}c@{}}\textbf{GraphRNN*}\\ \textbf{(w/ CBFS)}\end{tabular}} & \multicolumn{1}{c|}{\begin{tabular}[c]{@{}c@{}}\textbf{GEMS}\\ \textbf{$(\alpha=0)$}\end{tabular}} & \textbf{GEMS}  \\
\hline \hline
\multirow{3}{*}{MMD} & NSPDK* (x$10^3$) $\downarrow$ & 5.10 & \textbf{0.47} & 1.39 & 1.21 \\
& Node Label (x$10^4$) $\downarrow$ & 5.26 & \textbf{5.14} & 5.16 & 5.19\\
& Edge Label (x$10^2$) $\downarrow$ & 6.19 & 1.70 & \textbf{1.07} & 1.13 \\
\hline
\hline
\multirow{3}{*}{Node Metrics} & Count Reference & \multicolumn{4}{c|}{11.09} \\
\hhline{~|-|----}
& Count Predicted & 13.84 & 9.70 & 9.23 & 10.17 \\
& $(Obj)_{K}$ (x$10^2$) $\uparrow$ & 86.9 & 92.8 & 92.6 & \textbf{92.9}\\
\hline
\hline
\multirow{3}{*}{Edge Metrics} & Count Reference & \multicolumn{4}{c|}{5.01} \\
\hhline{~|-|----}
& Count Predicted & 11.86 & 2.74 & 6.84 & 7.45 \\
& MEP (x$10^2$) $\uparrow$ & 25.52 & 19.9 & 35.40 & \textbf{35.81} \\
\hline
\end{tabular}
\end{adjustbox}
\vspace{0.25cm}
\caption{Evaluation of different components of our method on Visual Genome: GraphRNN with $max\_prev\_node=6$ (GraphRNN*); GraphRNN* with Cluster-Aware Sequencing (GraphRNN* w/ CBFS); Our method without the use of external knowledge in the node loss (GEMS$(\alpha=0)$); The final model GEMS including all components.}
\label{table:ablationResults}
\end{table*}

\subsection{Evaluation Protocol}
Evaluation of generative models is a difficult task~\cite{theis2015note}. Current practice within the graph generation community is the use of Maximum Mean Discrepancy (MMD) as a way to characterise the performance of alternative models. Given two samples of graphs $\mathbf{G}_1=\{G_{11},G_{12},...,G_{1m}\}\sim\mathbb{G}$ and $\mathbf{G}_2=\{G_{21},G_{22},...,G_{2n}\}\sim\mathbb{G}$, the MMD between these two samples -- $MMD(f(\mathbf{G}_1), f(\mathbf{G}_2))$ -- is characterised by two factors: ($1$) a descriptor function, referred to as $f$, that returns a distribution of some chosen property over the set; and ($2$) a kernel function that computes the distance between the distributions. We consider three classes of descriptor functions capturing \textbf{Structural} (number of nodes, number of edges, node degree, clustering coefficient) and \textbf{Label} (node and edge types) properties of the graphs; as well as \textbf{Sub-Graph Similarities} (referred to NSPDK  from~\cite{costa2010fast}). 

The complexity of evaluating MMD is quadratic with respect to the number of samples in each set~\cite{JMLR:v13:gretton12a}. 
Computing MMD over subsets of the test set, also followed by~\cite{graphgen}, is one way to handle the computational cost of evaluation. Others~\cite{liao2019efficient,garg2021unconditional} propose alternative faster kernels to achieve the same end. The area of evaluation of graph generation models remains an open and active topic~\cite{obray2021evaluation}, future work will look into the effects of these design choices. In this work, we have followed the \cite{graphgen} and for Visual Genome we report the average value of MMD metrics calculated on $4$ independent splits of test set. For VRD we calculate MMD metrics on entire test set together. Our choice of kernel for MMD computation is the commonly used Gaussian kernel. In the results, we also report \textit{Novelty}, as defined by~\cite{graphgen}, which computes the fraction of expanded scene graphs that are not sub-graph isomorphic to graphs in the training set. In the next section, we describe other metrics customized to the domain of scene graph generation.

In addition, similar to \cite{scenegraphgen2021}, we evaluate the quality of images generated from the expanded scene graphs between our method and the baselines using traditional metrics to evaluate quality of synthesized images, namely, Frechet Inception Distance (FID) \cite{heusel2017gans}, Precision ($F_8$) and Recall ($F_{1\slash8}$) \cite{sajjadi2018assessing}, and Inception Score (IS) \cite{salimans2016improved}. The images are generated from the expanded scene graphs using pretrained models for Visual Genome dataset provided in sg2im \cite{johnson2018image} at a resolution of 64x64. The generated images are compared against Ground-truth images from Visual Genome dataset.

\subsubsection{Metrics for Scene Completion}
In this section, we introduce two new metrics to evaluate the output of scene graph generation methods. While we are utilizing them in a conditional setting, they are also valid for unconditional generation.

\textbf{Top-K Object Co-occurrence} $(Obj)_K$ The co-occurrence of a pair of objects in a set of graphs is calculated as the conditional probability of observing the pair in a scene graph given that one of the objects is present in the scene graph. We compare the co-occurrence of the $K$-most commonly observed pairs of objects in the test set with the co-occurrence of the corresponding pairs in the generated set of graphs as follows:\\
\begin{equation}
    (Obj)_{K}=1-\frac{1}{K}\,\smashoperator{\sum_{\substack{v_i,v_j\,\in\,\\top_K(P_{test})}}}\mid P_{test}[i,j]-P_{gen}[i,j]\mid
\end{equation}
Here, $P_{test}$ ($P_{gen}$) is a matrix such that entry $(i,j)$ is the co-occurrence of the pair of objects $(v_i,v_j)$ in the test set (generated set respectively).
In combination with the other metrics, $(Obj)_{K}$ rewards a model that generates graphs containing coherent objects.

\textbf{Modified Edge Precision (MEP)} is a metric inspired from modified n-gram precision~\cite{papineni2002bleu} popularly used in NLP.
\begin{align}
    MEP &= \underbrace{\min(1,exp^{(1-r/c)})}_{Brevity Penalty} \times \underbrace{\frac{\sum_{e \in G_{E} \cap (D_{E} \cup T_{E})} 1}{\sum_{e \in G_{E}} 1}}_{Edge Prec.}
\end{align}
Here, $G_E$, $D_E$ and $T_E$ refer to the set of directed edges present in the generated graph $G$, the training set and the test set respectively. The variable $r$ is the average number of edges in the reference graphs, taken to be the set of test graphs containing the seed graph $G_s$ for which $G$ is the expansion. $c$ is the number of edges in the expanded scene graph $G$. 
Note that this metric brings information orthogonal to the others, as shuffling the edge labels in a scene graph would yield the same score on the remaining metrics.

In addition, we use an alteration to the Neighbourhood Sub-graph Pairwise Distance Kernel (NSPDK) based MMD metric. NSPDK computes the distance between two graphs by matching pairs of sub-graphs with different radii $r$ and distances $d$. Since the Node label MMD already does a node-level comparison, we exclude $(r,d)=(0,0)$ and start from $(r,d)=(0,1)$ instead. The altered metric, referred to as NSPDK* in the results, better captures the contribution of larger sub-graphs. 

\begin{table}[t]
\centering
\begin{adjustbox}{ width=0.7\textwidth}
\renewcommand{\arraystretch}{1.3}
\begin{tabular}{|c|c|c|c|c|}
\hline 
 & \textbf{FID} $(\downarrow)$ & \textbf{Inception} $(\uparrow)$ & \textbf{Precision ($F_8$)} $(\uparrow)$ & \textbf{Recall ($F_{1\slash8}$)} $(\uparrow)$ \\
\hline
GraphRNN* & 173.72 & 4.49 $\pm$ 0.06 & 0.031 & 0.185 \\
\hline
SceneGraphGen & \textcolor{red}{\textbf{157.80}} & \textcolor{red}{\textbf{5.15 $\pm$ 0.135}} & \textcolor{blue}{\textbf{0.040}} & \textcolor{blue}{\textbf{0.235}} \\
\hline
GEMS & \textcolor{blue}{\textbf{160.65}} & \textcolor{blue}{\textbf{5.11 $\pm$ 0.1}} & \textcolor{red}{\textbf{0.045}} & \textcolor{red}{\textbf{0.240}} \\
\hline
\end{tabular}
\end{adjustbox}
\vspace{0.25cm}
\caption{Comparison of the quality of images generated using sg2im \cite{johnson2018image} on the expanded scene graphs by different models for the same seed scene graph on Visual Genome dataset. \textbf{Note:} \textcolor{red}{\textbf{red}} represents best and \textcolor{blue}{\textbf{blue}} represents second best scores.}
\label{table:diversityResults}
\end{table}

\begin{figure*}
\begin{center}
\includegraphics[width=\linewidth]{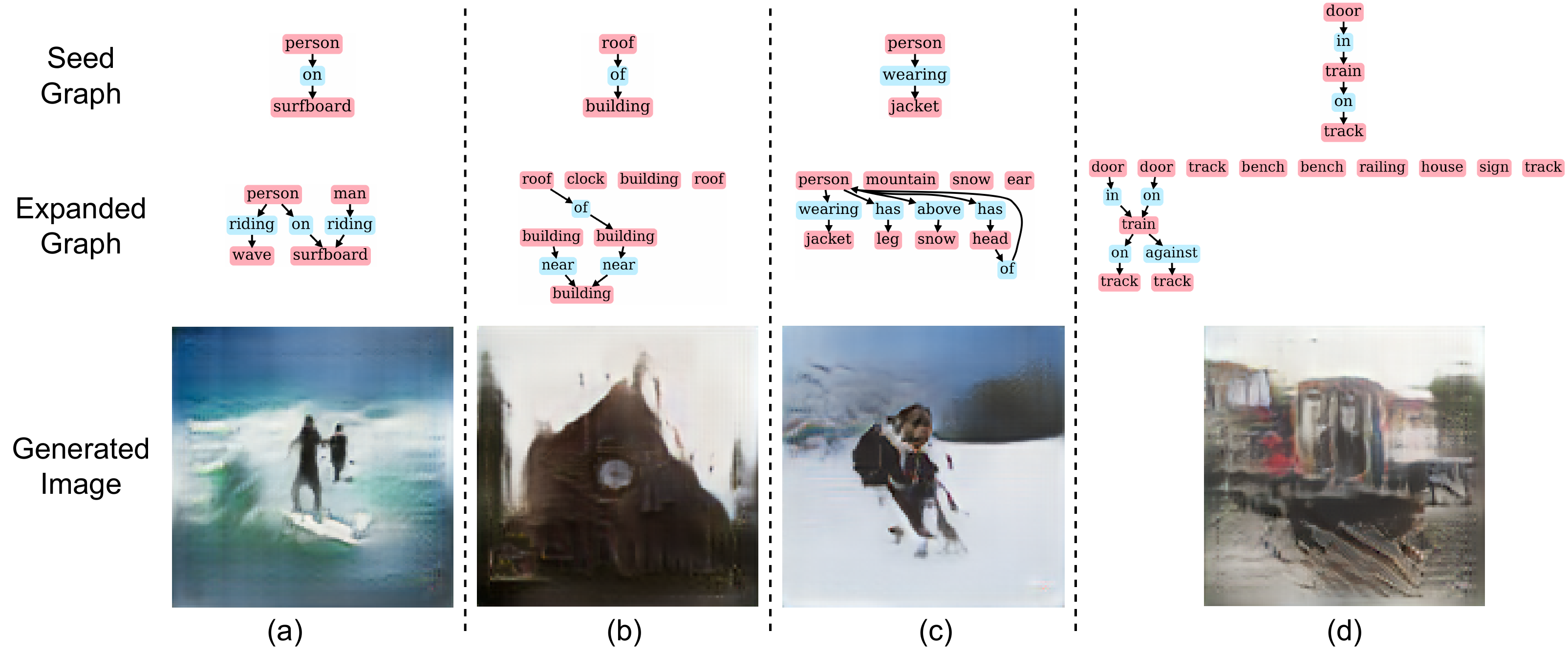}
\end{center}
    \vspace{-0.5cm}
  \caption{Examples of images generated by sg2im \cite{johnson2018image} using expanded scene graphs from seed graphs using our method (GEMS).}
\label{fig:images}
\end{figure*}

\subsection{Main Results}
Table 1 shows the comparison of graphs generated by GEMS and baselines methods in different metrics. For Visual Genome our model outperforms the GraphRNN based baseline methods by significant margin on all metrics demonstrating graphs generate by our method are more meaningful and more closely resemble the observed scene graph distribution. For VRD our method shows comparable results to GraphRNN*, but outperforms GraphRNN in almost all metrics. We also provide SceneGraphGen results for comparison, and the key contributor for SceneGraphGen's performance are the architectural advances that we do not use in our work. Still our method outperforms SceneGraphGen in terms Node Label and Degree MMD metrics for Visual Genome dataset, indication better node predictions in the expanded scene graphs for Visual Genome. 
Figure~\ref{fig:baselines} shows a qualitative comparison of graphs generated by GEMS and GraphRNN*, and SceneGraphGen. It can we seen that our model expands seed graphs with diverse nodes that occur together in natural scenes. The relationships amongst nodes are also meaningful and appropriate. However, GraphRNN* produces graphs with irrelevant relationships between nodes (e.g. $surfboard-wearing-arm$ and $snow-has-pole$).

Table 3 shows a comparison of the quality of images generated using sg2im  \cite{johnson2018image} on the expanded scene graphs by our model and the baselines for the same seed graph on Visual Genome dataset. Our model performs better in terms of Precision and Recall compared to all the baselines. For FID and Inception  however, ScenegraphGen slightly outperforms our model (GEMS).
Figure~\ref{fig:images} shows diverse seed graphs, the corresponding expanded scene graphs generated by our method (GEMS), and the images generated by sg2im \cite{johnson2018image} using the expanded scene graphs. We can see that from an abstract seed graphs like $roof-on-building$ or $person-wearing-jacket$, it can generate complete and meaningful scenes like building with roofs having a large clock at the front or a person wearing jacket on a snowy mountain.

\subsection{Ablation Study}
\subsubsection{Cluster-Aware BFS} 
\begin{figure}[!t]
\begin{center}
\includegraphics[width=0.9\linewidth]{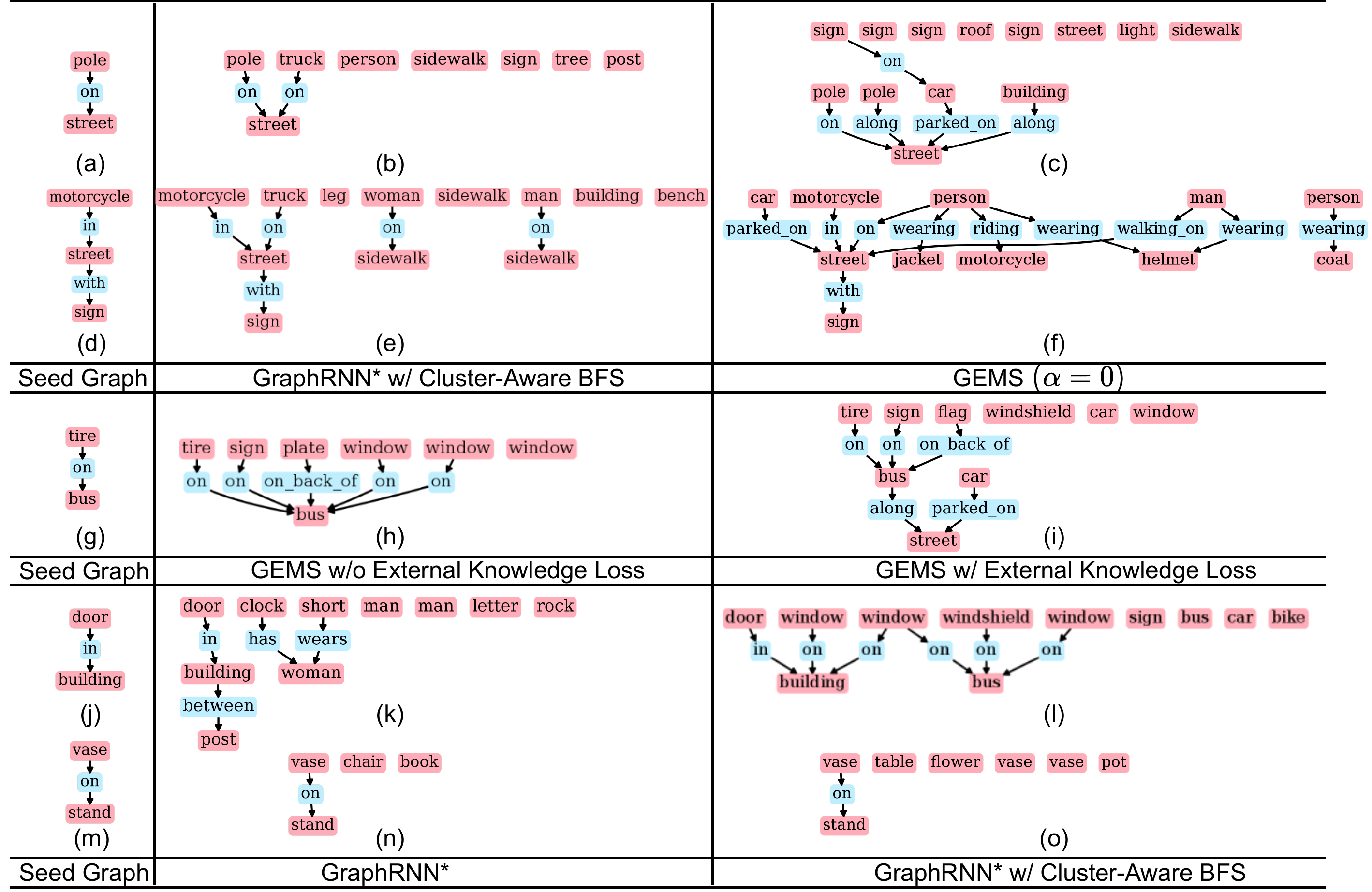}
\end{center}
    \vspace{-0.5cm}
\caption{Figure shows the behavior brought by each component of our model in isolation (Middle and Right column results are without and with using a component respectively). $1^{st}$ two rows show the benefits of Subject-Object addition to edge prediction model, the use of GloVe embeddings and class-balancing edge loss. $3^{rd}$ depicts advantage of using external-knowledge for node prediction. The last two rows show the advantage of Cluster-Aware BFS.}
\label{fig:ablation}
\end{figure}

From Table 2, it is observed that Cluster-Aware BFS drastically improves the performance on NSPDK* and $(Obj)_{k}$ indicating that generated graphs respect cluster (and even pair) of objects that occur together in observed set of graphs in the of training set. Fig. \ref{fig:ablation} (last 2 rows) shows qualitative example for benefits of Cluster-Aware BFS. (l) adds the cluster - $window, windshield, bus$ to the seed graph because these 3 objects would occur together in observed scenes also. In (o) $flower, table$ are added to seed graph $vase-on-stand$ as $flower$ and $table$ occur together with $vase$ and $stand$ respectively.

\vspace{-0.25cm}
\subsubsection{Subject-Object Context \& Class-Balancing Loss}
Adding subject-object context to EdgeRNN for edge prediction enables the model to predict only meaningful relationships which are observed in training distribution between the given subject and object. Additionally, Class-balancing Loss is required to tackle the skewness in our scene graph dataset such that model produces diverse set on relationships (and not just dominant ones like $on$ and $has$). Both these components are designed towards improving quality of predicted edges. Table 2 Edge Metrics show significant improvement in MEP and edge count compared to models without these components, validating our hypothesis. Fig. 2 Top 2 rows, although (b), (c) and (e), (f) contain relatively similar objects in (b) and (e), the model without subject-object context and class-balancing loss produces graphs with only $on$ and $with$ relations (that occur most of the times in Visual Genome), however (c) and (f) contains diverse edges like $parked\_on$, $wearing$, $riding$, $walking\_on$, $walking$ in addition to $on$ and $with$.
\vspace{-0.25cm}
\subsubsection{External Knowledge Loss}
Adding external knowledge loss helps in addition of relatively similar but diverse nodes and not repeated node labels. This evident from Fig. \ref{fig:ablation} middle coloumn. In (h) the model adds $window$ several times with $bus$, but in (i) the models adds both $window$ and $windshied$ (different node labels but similar in terms of glove embeddings used in external-knowledge loss). From Table 2 also it is seen that adding external-knowledge loss imporves $(Obj)_k$.

\begin{figure}
\begin{center}
\includegraphics[width=0.7\linewidth]{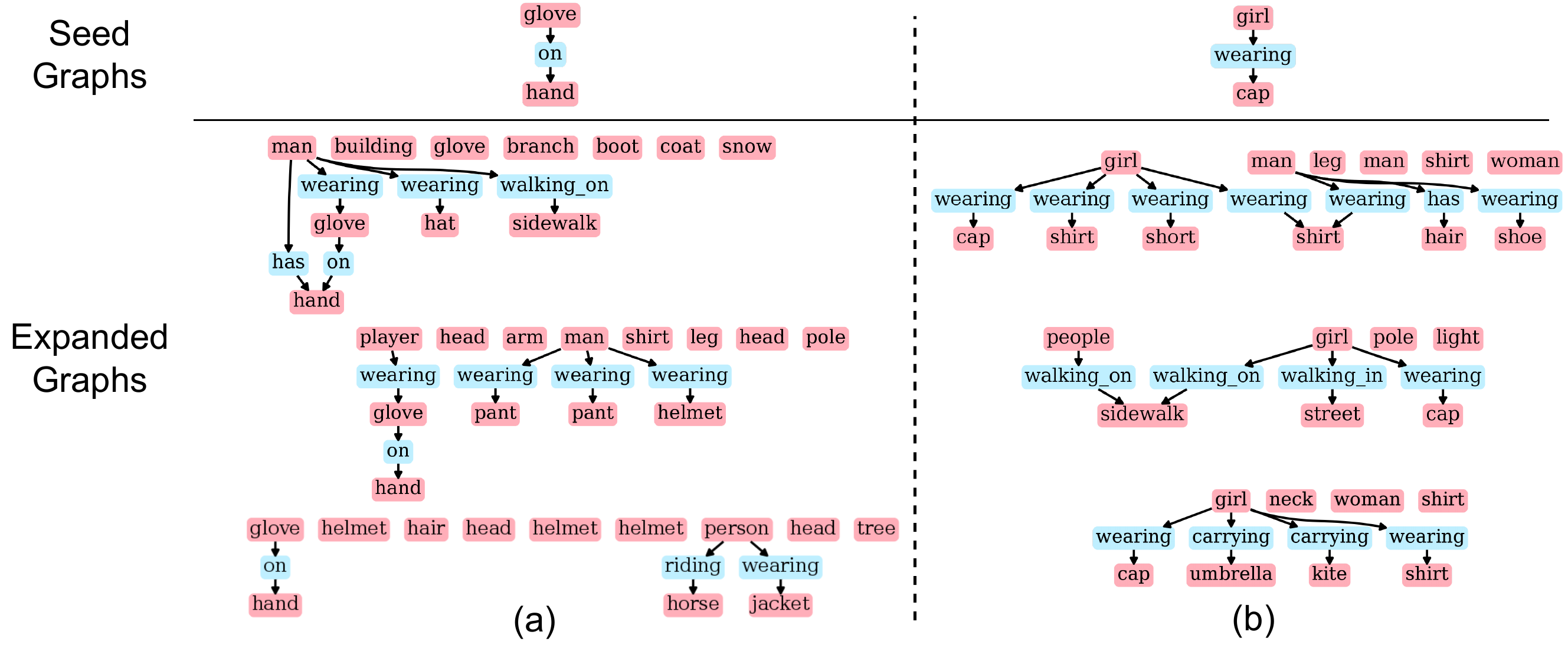}
\end{center}
    \vspace{-0.5cm}
\caption{Three different scene graph expansions produced by our model (GEMS) for the same input seed graph. GEMS not only adds diverse objects to the seed graph but also generates diverse visual scenarios.}
\label{fig:diversity}
\vspace{-0.3cm}
\end{figure}

\subsubsection{Multiple Outputs}

At every step of GEMS, a node is first sampled from the multinomial distribution over node labels (output of $f_{node}$), and edges are added from previous sampled nodes to the new one (by $f_{edge}$). This process continues until an end-of-sequence node token is obtained. Our model can be used to generate $M$ alternative expansions of the same seed scene graph by invoking the sampling process multiple times with different seeds -- Figure~\ref{fig:diversity} show provides qualitative examples. 

\section{Discussion and Conclusion}
In this paper, we considered the novel task of \textit{scene graph expansion} -- given an input seed scene graph, we enhance it by the addition of objects and relationships. The output, representing a more complex scene, is expected to respect observed co-occurrence patterns between objects and their relationships, while being novel with respect to the training set. Doing so automatically is enabled by the use of generative models of graphs, which provide a scalable mechanism to model real world graphs in multiple domains. The training of our autoregressive model is guided by external knowledge captured via embeddings from the linguistic domain.

Our extensive experimental section illustrated that the standard MMD based evaluation does not highlight all behavioral characteristics of the models. In particular, we confirm the observation made by others that while models achieve satisfactory results on object-centric prediction tasks, modeling relationships is harder~\cite{wang2019exploring}. We propose new metrics specifically focussed on this aspect, and compare our models and baselines on the new metrics. However, evaluation of conditional and unconditional generation of scene graphs remains challenging. In particular, the design of metrics that capture the semantic plausibility of a generated scene graph is an important future direction. As with other generative models, there is a balancing act of staying within the support offered by the training set (captured by MMD metrics) while producing unseen output. Our metrics and experiments primarily focused on the behavior of the models with respect to structural graph properties. 

For quantitative evaluation, we define a diversity metric, from a set of $M(=3)$ expansions, we exclude the expansion that is sub-graph isomorphic to one of the other expansions, and compute the percentage that remains. The results provided in Table~\ref{table:baselineResults} suggest a bias-variance trade-off. Across datasets, GraphRNN* has higher diversity, indicating that the predicted distribution over node labels at every step is flatter. Encouraging our GEMS model to produce diverse variations, while still respecting the training set distribution (captured by other metrics like NSPDK and MEP) remains a topic for future work.


%
%
\bibliographystyle{splncs04}
\bibliography{egbib}

\appendix

\section{Training and Testing Algorithms}
In this section, we present the pseudo code for training \ref{alg:training} and inference \ref{alg:inference}. We use the notation as described in our main paper under problem description. $f_{node2edge}$ is an additional neural network (multi-layer perceptron) that is used to transfer information from $f_{node}$ to $f_{edge}$, i.e. the hidden state of $f_{edge}$ is initialized from the hidden state of $f_{node}$ using this transformation. 

\section{Selecting value of $max\_prev\_node$}

\begin{figure}
\begin{center}
\includegraphics[width=0.9\linewidth]{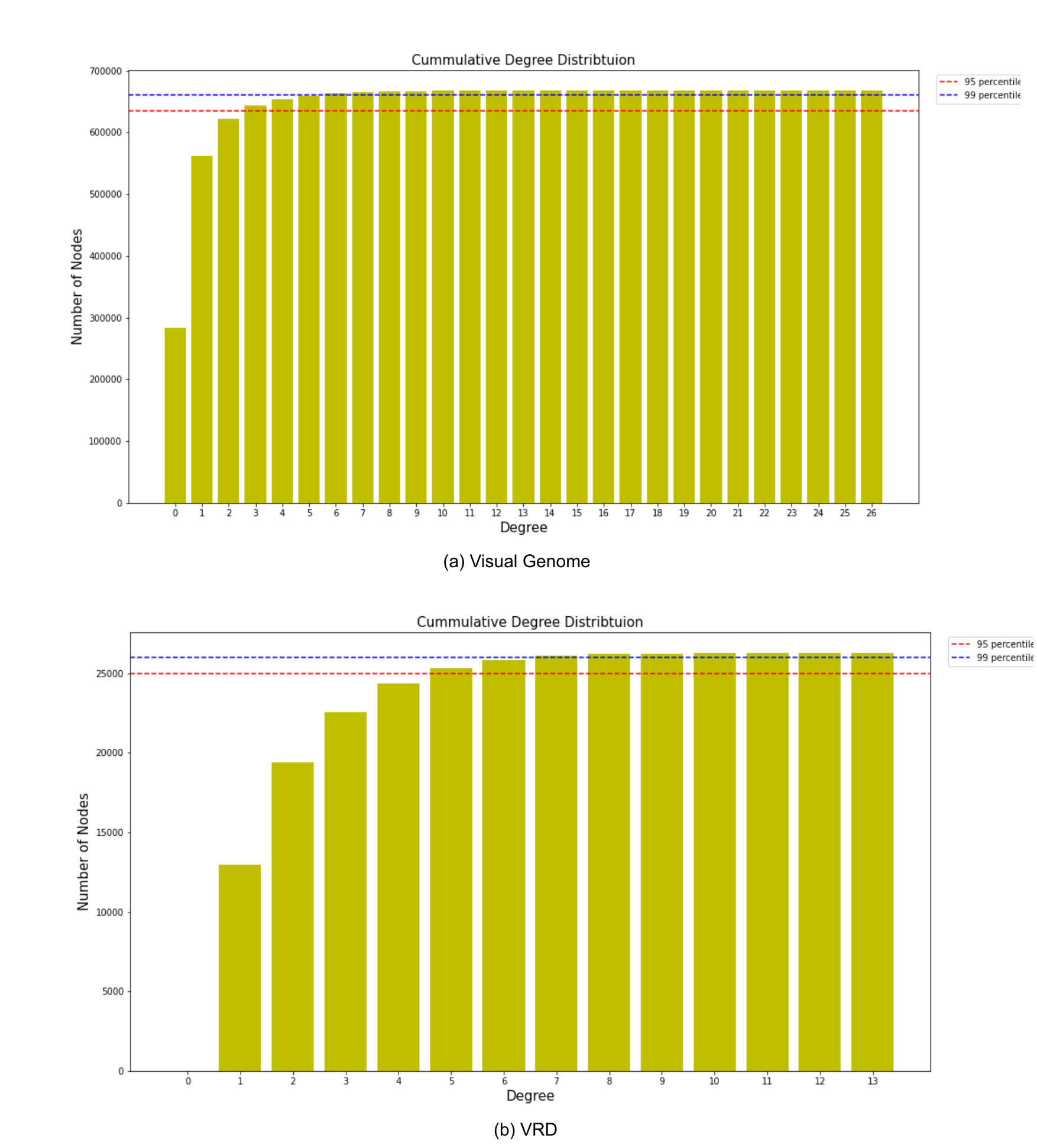}
\end{center}
    \vspace{-0.25cm}
  \caption{Cumulative degree distribution for the two datasets, Visual-Genome and VRD.}
\label{fig:max-prev-node}
\vspace{-0.5cm}
\end{figure}

Our method GEMS (and the baseline adapted from GraphRNN) requires the EdgeRNN to predict edges between the newly generated node $v_i$ and the previous nodes in the graph sequence. The number $k$, referred to as $max\_prev\_node$, controls how many of the previously chosen nodes to take into account for predictions, and is a hyperparameter of the underlying method. If the value of $k$ is set to a very large number, the model generates edges much larger in number than that is expected in training distribution (see count predicted and count reference rows in Table 1 in the main paper). If $k$ is calculated to using the method described by GraphRNN \cite{graphrnn}, we obtain $k=37$ for the Visual Genome dataset. This leads to generated graphs containing large number of edges ($29.53$) compared to reference number of edges in training set ($11.09$), and most of the edges are either degenerate or describe irrelevant relationships between the subject and object. To address this problem we compute the value of $k$ as the maximum degree (out-degree + in-degree) of a node in a graph such that $99\%$ of nodes in all graphs in the training set have degree less than or equal to $k$.  
By setting $max\_prev\_node$ to $k$ chosen in this manner, we lose only a few distant relationships between nodes in a graph. However, the use of BFS on maximal connected subgraphs in our Cluster-Aware BFS algorithm ensures that the objects that are closely related in the observed set of graphs appear together in sequence, further reducing the chances of not retaining edges between nodes which are related. As observed from Table 1 in main paper GraphRNN*, which has $max\_prev\_node = 6$ for Visual Genome, replicates the count of predicted edges in the generated graphs much more faithfully that GraphRNN. Hence, we gain efficiency and improve performance at the cost of a negligible number of relationships. From the plots of the cumulative degree distribution in Figure~\ref{fig:max-prev-node}, we observe that $k$ is equal to $6$ for Visual Genome and it is equal to $7$ for VRD.

\begin{algorithm}[t]
\caption{GEMS Training Algorithm}\label{alg:training}
\begin{algorithmic}
\small
\REQUIRE Dataset of Graphs $\mathbb{G} = \{G_1,G_2,\dots,G_n\}$
\ENSURE Learned functions $f_{node}, f_{edge}, f_{node2edge}$
\STATE Initialize $f_{node}, f_{edge}, f_{node2edge}$
\FOR{epoch in epochs}
    \FOR{$G \in \mathbb{G}$}
        \STATE $S \gets S^{\pi}(G)$
        \STATE $S_0, h_{node}^0 \gets SOS, h_{init}$
        \FOR[{$n_G +1$ for EOS token}]{$i \in 1\dots n_G+1$}
            \STATE $p_{\hat{v}_i}, h_{node}^i \gets f_{node}(S_{i-1}, h_{node}^{i-1})$
            \STATE $q_i \gets \min_q KL(q,p_{\hat{v}_i}) -\mathbb{E}_{v\sim q}(f(v,v_i))$
            \STATE $start \gets max(1,i-k)$\COMMENT{$k$ is max\_prev\_node}
            \STATE $h_{edge}^{(start-1)} \gets f_{node2edge}(h_{node}^i)$ \COMMENT{ Initialize hidden state of $f_{edge}$}
            \STATE $L_{edge}^i \gets 0$
            \FOR{$j \in start\dots i-1$}
                \STATE $p_{\hat{e}_{i,j}}, h_{edge}^{temp} \gets f_{edge}(v_i, v_j, h_{edge}^{j-1})$
                \STATE $p_{\hat{e}_{j,i}}, h_{edge}^{j} \gets f_{edge}(v_j, v_i, h_{edge}^{temp})$
                \STATE $L_{edge}^i \gets L_{edge}^i + l_{edge}(p_{e_{i,j}}, p_{\hat{e}_{i,j}})$
                \STATE $L_{edge}^i \gets L_{edge}^i + l_{edge}(p_{e_{j,i}}, p_{\hat{e}_{j,i}})$
            \ENDFOR
            \STATE $L_{node}^i \gets l_{node}(p_{v_i}, p_{\hat{v}_i})$
        \ENDFOR
        \STATE $\mathcal{L}(G) \gets L_{node}^{n_G+1}+\displaystyle\sum_{i \in 1\dots n_G} L_{node}^i + L_{edge}^i$
    \ENDFOR
    \STATE {\scriptsize Back-propagate loss and update weights of} $f_{node}, f_{edge}, f_{node2edge}$
\ENDFOR \COMMENT{typically when validation loss is minimized}
\end{algorithmic}
\end{algorithm}

\begin{algorithm}[t]
\caption{GEMS Inference Algorithm}\label{alg:inference}
\begin{algorithmic}
\small
\REQUIRE Seed graph $G_s$; $f_{node}, f_{edge}, f_{node2edge}$
\ENSURE Expanded graph $\hat{G}_s$
\STATE $S \gets S^{\pi}(G_s)$; $\hat{S}_0, h_{node}^{0} \gets SOS, h_{init}$
\FOR[Conditioning on the input]{$i \in 1\dots n_{G_s}$}
    \STATE $p_{\hat{v}_i}, h_{node}^{i} \gets f_{node}(\hat{S}_{i-1}, h_{node}^{i-1})$
    \STATE $\hat{S}_i \gets S_i$
\ENDFOR
\REPEAT[Expanding the seed graph]
    \STATE $i \gets i+1$
    \STATE $p_{\hat{v}_i}, h_{node}^{i} \gets f_{node}(\hat{S}_{i-1}, h_{node}^{i-1})$
    \STATE $\hat{v}_i \sim p_{\hat{v}_i}$
    \STATE $start \gets max(1,i-k)$\COMMENT{$k$ is max\_prev\_node}
    \STATE $h_{edge}^{(start-1)} \gets f_{node2edge}(h_{node}^i)$
    \FOR{$j \in start\dots i-1$}
        \STATE $p_{\hat{e}_{i,j}}, h_{edge}^{temp} \gets f_{edge}(\hat{v}_i, \hat{v}_j, h_{edge}^{j-1})$
        \STATE $p_{\hat{e}_{j,i}}, h_{edge}^{j} \gets f_{edge}(\hat{v}_j, \hat{v}_i, h_{edge}^{temp})$
        \STATE $\hat{e}_{i,j} \sim p_{\hat{e}_{i,j}}$, $\hat{e}_{j,i} \sim p_{\hat{e}_{j,i}}$
        \STATE $\hat{\mathcal{E}}_{i,j} \gets (\hat{e}_{i,j}, \hat{e}_{j,i})$
    \ENDFOR
    \IF{$\hat{v}_i$ is not EOS}
        \STATE $\hat{S}_i \gets (\hat{v}_i, \hat{\mathcal{E}}_i)$
    \ENDIF
\UNTIL $\hat{v}_i$ is EOS
\STATE Convert $\hat{S}$ into graph $\hat{G}_s$

\end{algorithmic}
\end{algorithm}

\section{GraphRNN \cite{graphrnn} for Scene Graph Expansion (Baseline)}
We use GraphRNN \cite{graphrnn} as the baseline to compare against our method, GEMS, for the task of scene graph expansion. GraphRNN is an auto-regressive model that converts a graph to a sequence, and then successively predicts nodes and edges according to the sequence. The sequencing method used in their work is BFS starting from a randomly chosen node in the graph. This sequencing method was possible because the graphs which they work on are connected graphs, for which there is a valid BFS tracersal. However, scene graphs used in our task (both from Visual Genome \cite{visualgenome} and VRD \cite{vrd-and-language-priors}) tend to have disconnected components. Additionally, the scene graphs that we work with, contain bi-directional edges between nodes (the datasets that GraphRNN was evaluated on contain only undirected or unidirectional edges). To use their sequencing method we convert the disconnected scene graphs to connected scene graphs. More specifically, we add a dummy node labelled $image$ to each of the graphs, and connect this $image$ node with all other nodes ($v_i$) in the graphs using bidirectional dummy edges $in\_image$ (from $v_i$ to $image$) and $contains$ (from $image$ to $v_i$) and then perform BFS starting from any node in the graph. At inference time we generate the graph using these dummy nodes and edges, and then remove the dummy nodes and edges for evaluation. We modify the original GraphRNN code to predict bidirectional edges sequentially, i.e., for a newly generated node $v_i$ and a previously generated node $v_j$ in the graph, the EdgeRNN of GraphRNN first predicts the edge from $v_i$ to $v_j$ and then the edge from $v_j$ to $v_i$.

\section{SceneGraphGen for Scene Graph Expansion (Baseline)}
We compare our method GEMS to SceneGraphGen~\cite{scenegraphgen2021}, which was primarily designed for unconditional generation of scene graphs, for the task of scene graph expansion. Since SceneGraphGen also predicts nodes and the corresponding edges from previously generated nodes in an auto-regressive manner, they require an ordering method for converting graph to a sequence. The authors suggest a random ordering to flatten the graph into a sequence, and show that this performs best compared to other ordering methods -- we therefore use the same ordering strategy while evaluating SceneGraphGen. For both Visual Genome and VRD we use 
$num\_node\_categories = 153$, $num\_edge\_categories = 55$ and
$num\_node\_categories = 102$, $num\_edge\_categories = 73$ respectively, based on the dataset statistics. The other parameters for the models were retained as provided in the official github repository of SceneGraphGen
\\
\href{https://scenegraphgen.github.io/}{https://scenegraphgen.github.io/}. (Code is currently not available).

\section{Common Sense Knowledge Graph Embeddings}

\begin{table*}[t]
\centering
\begin{adjustbox}{ width=0.9\textwidth}
\renewcommand{\arraystretch}{1.3}
\begin{tabular}{|c|c|c|c|c|c|}
\hline 
 & & \multicolumn{4}{|l|}{\hspace{3.5cm}\textbf{Visual Genome}} \\
\hhline{~~|----}
 & & \textbf{GraphRNN} & \textbf{GraphRNN*} & \textbf{GEMS (GloVe)} & \textbf{GEMS (cskg)} \\
\hline \hline
\multirow{6}{*}{MMD} & Degree (x$10^2$) $\downarrow$ & 47.47 & 16.44 & \textbf{2.11} & 16.71 \\
& Clustering (x$10^2$) $\downarrow$ & 18.63 & 4.05 & \textbf{0.86} & 1.31 \\
& NSPDK* (x$10^3$) $\downarrow$ & 22.6 & 5.10 & \textbf{1.21} & 2.71 \\
& Node Label (x$10^4$) $\downarrow$ & 5.44 & 5.26 & 5.19 & \textbf{5.18} \\
& Edge Label (x$10^2$) $\downarrow$ & 22.38 & 6.19 & \textbf{1.13} & 3.34 \\
\hline
\hline
\multirow{3}{*}{Node Metrics} & Count Reference & \multicolumn{4}{|c|}{11.09} \\
\hhline{~|-|----|}
& Count Predicted & 29.53 & 13.84 & 10.17 & 10.97 \\
& $(Obj)_{K}$ (x$10^2$) $\uparrow$ & 83.7 & 86.9 & \textbf{92.9} & 85.9 \\
\hline
\hline
\multirow{3}{*}{Edge Metrics} & Count Reference & \multicolumn{4}{|c|}{5.01} \\
\hhline{~|-|----|}
& Count Predicted & 57.95 & 11.86 & 7.45 & 13.50 \\
& MEP (x$10^2$) $\uparrow$ & 22.4 & 24.52 & \textbf{35.81} & 35.12 \\
\hline
\hline
\multicolumn{2}{|c|}{Novelty (x$10^2$) $\uparrow$} & 12.26 & 57.59 & \textbf{75.75} & 68.96 \\
\hline
\end{tabular}
\end{adjustbox}
\vspace{0.25cm}
\caption{Comparison of 2 variants of our method, GEMS with glove embeddings and common-sense knowledge graph (cskg) \cite{BridgingKGZareian2020} in input and external knowledge loss with the baselines of GraphRNN \cite{graphrnn} and GraphRNN* on Visual Genome dataset. For all MMD based metrics, lower is better $(\downarrow)$. For the rest of the metrics, larger is better $(\uparrow)$. GraphRNN* refers to GraphRNN with $max\_prev\_node=6$ (Visual Genome) }
\label{table:baselineResults}
\end{table*}

We wanted to design a method that is agnostic to the type of embeddings (both at the input level as well as in the loss function as external knowledge) used to represent nodes and relationships. Hence, we experiment with the use of embeddings derived from the Common-Sense Knowledge Graph (CSKG, proposed in ~\cite{BridgingKGZareian2020}), whose vocabulary is the same as that of objects and relations in Visual Genome dataset.  To obtain the knowledge graph embeddings, we train ComplEx model~\cite{trouillon2016complex} using the OpenIE framework ~\cite{han2018openke} on the CSKG. Our hypothesis was that the nature of the information carried in the knowledge graph and word embeddings would be very different (co-occurrence in graph context, versus in unstructured text). However, we did not observe significant differences in the end results obtained using CSKG embeddings compared to GloVe embeddings \cite{pennington2014glove}. Table 1 shows comparison in terms of metric values computed on graphs generated by GEMS (using glove and common-sense knowledge graph embeddings) with GraphRNN and GraphRNN*.

\section{Additional Proposed Metrics}
In this section we describe the novel metrics that we have proposed for the evaluation of expanded scene graphs. We also provide the metric values computed on baselines GraphRNN, GraphRNN* and SceneGraphGen, and our model (GEMS) for Visual Genome \cite{visualgenome} and VRD \cite{vrd-and-language-priors} datasets.


\begin{table*}[t]
\centering
\begin{adjustbox}{ width=0.9\textwidth}
\renewcommand{\arraystretch}{1.3}
\begin{tabular}{|c|c|c|c|c|c|c|c|c|c|}
\hline 
 & & \multicolumn{4}{|l|}{\hspace{2cm}\textbf{Visual Genome}} & \multicolumn{4}{|l|}{\hspace{2.8cm}\textbf{VRD}} \\
\hhline{~~|--------}
 & & \textbf{GraphRNN} & \textbf{GraphRNN*} & \textbf{SceneGraphGen}  & \textbf{GEMS} & \textbf{GraphRNN} & \textbf{GraphRNN*} & \textbf{SceneGraphGen} & \textbf{GEMS} \\
\hline \hline

\multirow{2}{*}{Edge Metrics} & $(Trip)_{K}$ (x$10^2$) $\uparrow$ & 34.6 & 44.7 & \textcolor{red}{\textbf{72.9}} & \textcolor{blue}{\textbf{52.8}} & 35.1 & 37.3 & \textcolor{red}{\textbf{43.9}} & \textcolor{blue}{\textbf{38.1}} \\
 & ZSEP (x$10^2$) $\uparrow$ & 3.19 & \textcolor{blue}{\textbf{3.41}} & \textcolor{red}{\textbf{10.1}} & 3.14 & 2.76 & 2.93 & \textcolor{red}{\textbf{6.2}} & \textcolor{blue}{\textbf{3.18}} \\
\hline
\end{tabular}
\end{adjustbox}
\vspace{0.25cm}
\caption{Comparison of our method (GEMS) with the baselines of GraphRNN \cite{graphrnn}, GraphRNN* and SceneGraphGen \cite{scenegraphgen2021} on Visual Genome \cite{visualgenome} and VRD \cite{vrd-and-language-priors} datasets on additional metrics, Top-K  Triplet  Co-occurrence ($(Trip)_{K}$) and Zero-shot Edge Precision (ZSEP) . For both the metrics, upper is better $(\uparrow)$. GraphRNN* refers to GraphRNN with $max\_prev\_node=6$ and 7 for Visual Genome and VRD respectively in Table 2. \textbf{Note:} \textcolor{red}{\textbf{red}} represents best and \textcolor{blue}{\textbf{blue}} represents second best scores.}
\label{table:baselineResults}
\end{table*}

\subsubsection{Top-K Triplet Co-occurrence} $(Trip)_k$ The co-occurrence of a triple (Subject, Predicate, Object) in a set of graphs is calculated as the conditional probability of the predicate connecting the subject, object pair given that the pair is present. 
We compare the co-occurrence of the K-most commonly observed triples in the test set with the co-occurrence of the corresponding triples in the generated set of graphs as follows:
\begin{equation}
    (Trip)_{K}=1-\frac{1}{K}\,\smashoperator{\sum_{\substack{v_i,e_j,v_k\,\in\,\\top_k(P_{test})}}}\mid P_{test}[i,j,k]-P_{gen}[i,j,k]\mid
\end{equation}
Here, $P_{test}$ ($P_{gen}$) is a matrix such that entry $(i,j,k)$ is the co-occurrence of the triple $(v_i,e_j,v_k)$ in the test set (generated set respectively).
In combination with the other metrics, $(Trip)_{K}$ rewards a model that generates graphs containing coherent relations between two objects with similar probabilistic distribution as observed in training set. Note that a trigram MMD metric that compares the triplet distributions in the real and generated sets would achieve the same purpose. However, given the computational complexity of computing such an MMD-based metric, we propose the metric as defined above.

\subsubsection{Zero-Shot Edge Precision} $ZSEP$ This is a metric inspired from zero shot learning \cite{chang2008datalessclass,larochelle2008zero} to compute the relevance of the novel edges being generated by the model. \cite{vrd-and-language-priors} proposed a metric along these lines for zero shot visual relationship detection, which was based on recall. Our proposed metric computes the fraction of novel edges generated by the model which are present in the test data. The presence of a relationship in the test data implies that it is realistic, hence this fraction is a surrogate measure of how well the model learns about unseen relationships by leveraging similar relationships which it has already seen in the training distribution.
\begin{align}
    ZSEP = \frac{\sum_{e \in (G_E \backslash D_E) \cap T_E} 1}{\sum_{e \in (G_E \backslash D_E)} 1}
\end{align}
Here, $G_E$, $D_E$ and $T_E$ refer to the set of directed edges present in the generated graph $G$, the training set and the test set respectively. 


\section{Additional Qualitative Results}

\begin{figure*}
\begin{center}
\includegraphics[width=\linewidth]{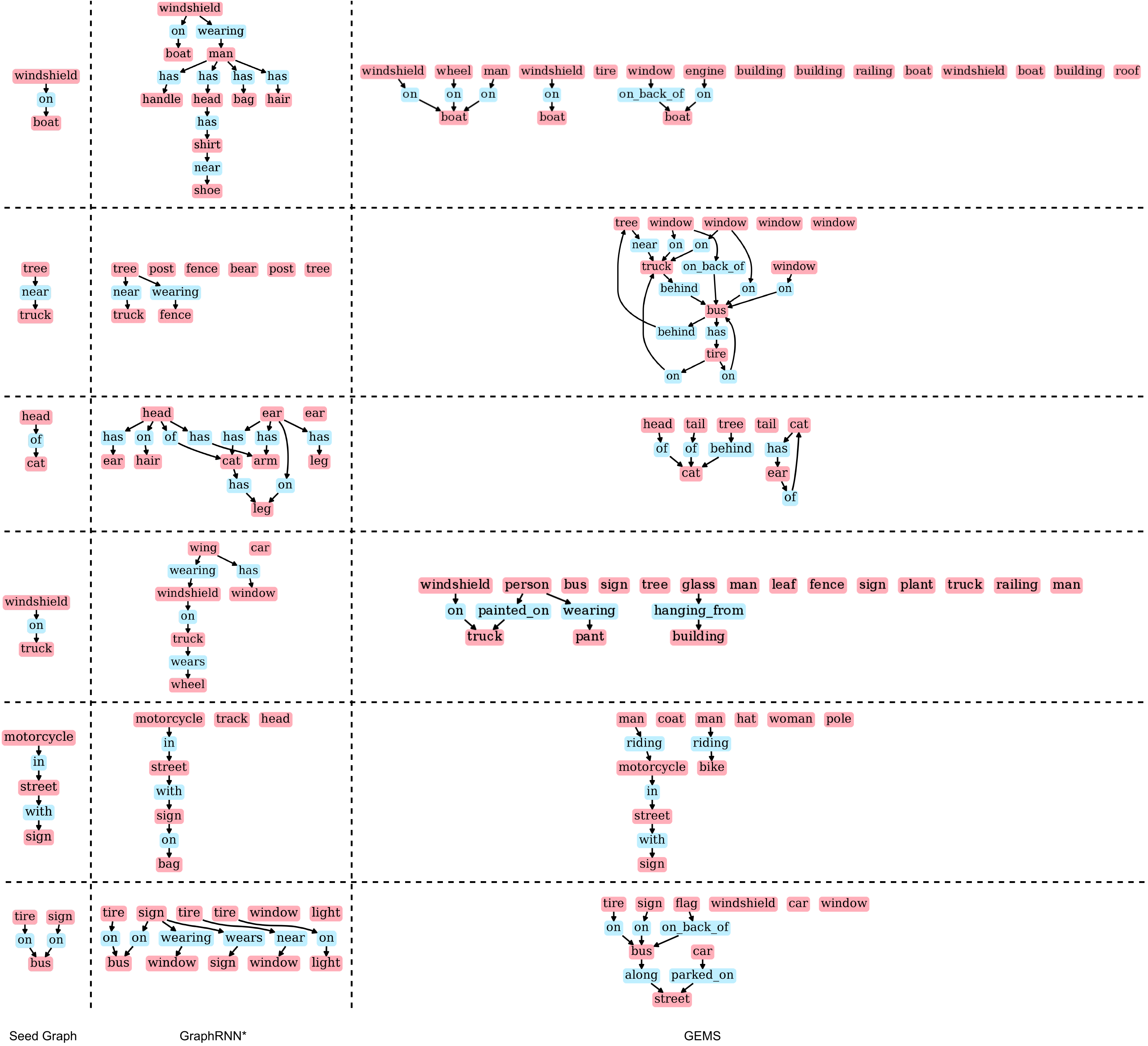}
\end{center}
  \caption{Additional comparison of expanded graphs generated by our model (GEMS) v/s baseline GraphRNN* (GraphRNN with $max\_prev\_node=6$) on Visual Genome seed graphs. 
  }
\label{fig:baselines1}
\end{figure*}

\begin{figure*}
\begin{center}
\includegraphics[width=\linewidth]{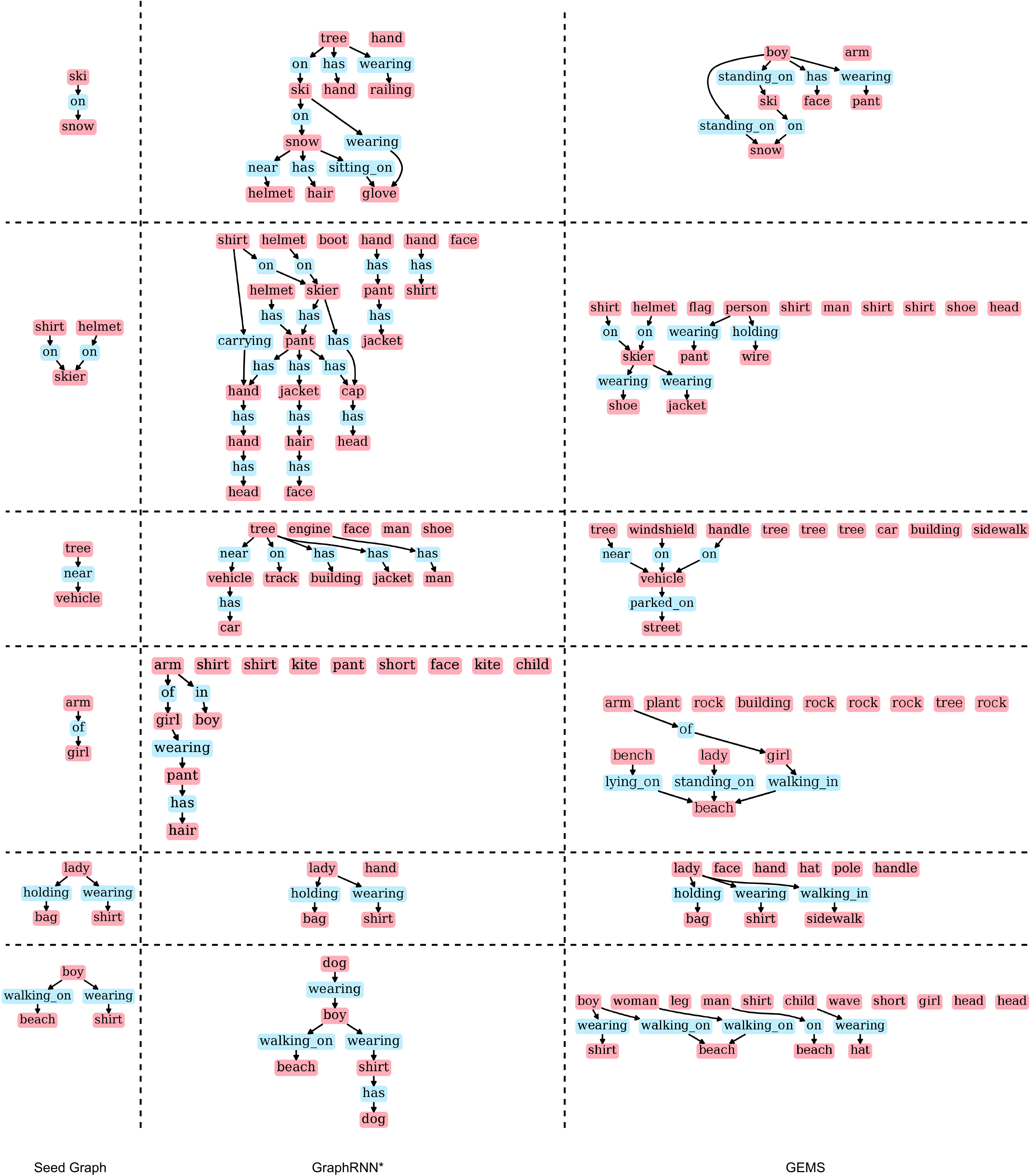}
\end{center}
  \caption{Additional comparison of expanded graphs generated by our model (GEMS) v/s baseline GraphRNN* (GraphRNN with $max\_prev\_node=6$) on Visual Genome seed graphs. 
  }
\label{fig:baselines2}
\end{figure*}

\begin{figure*}
\begin{center}
\includegraphics[width=\linewidth]{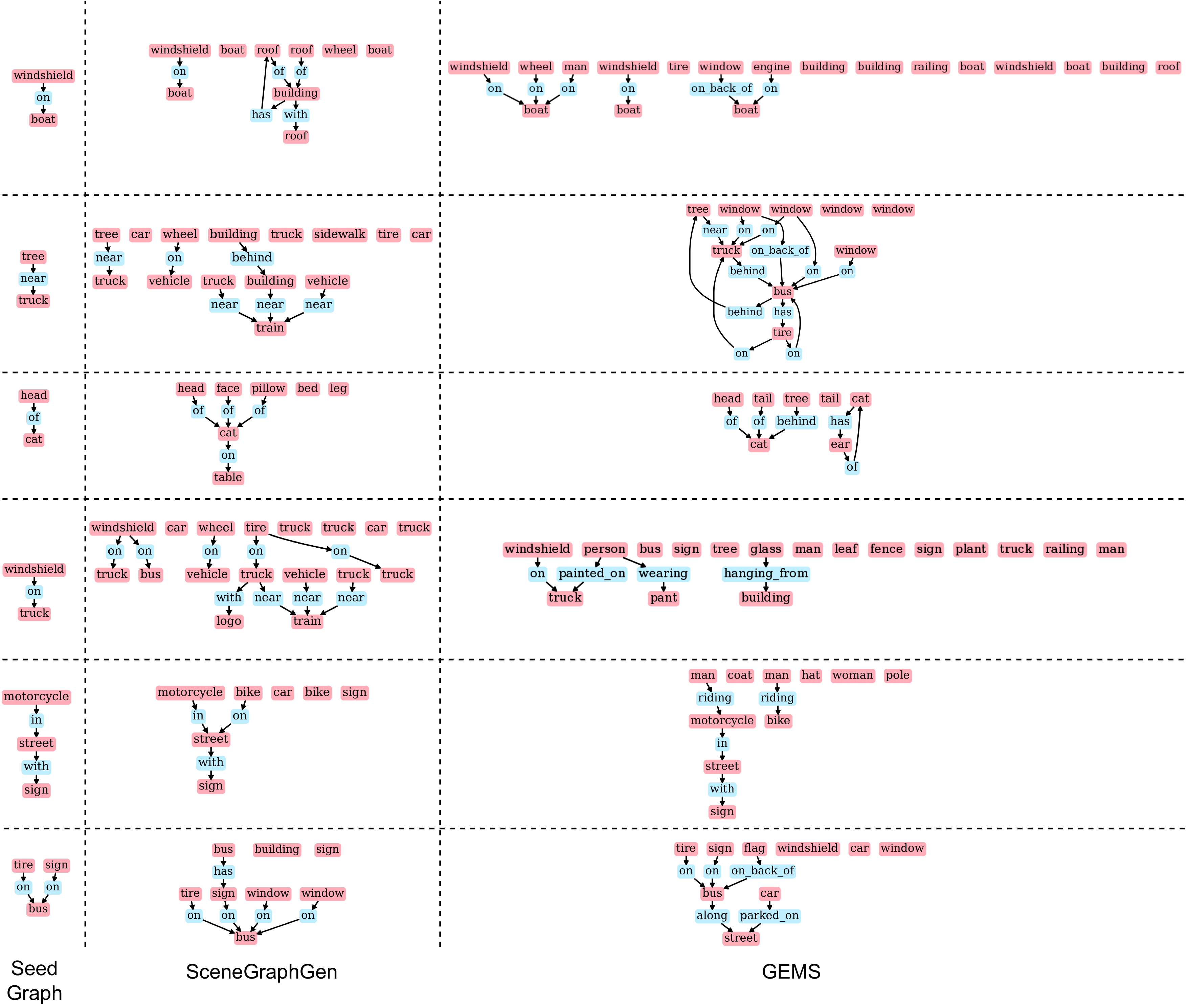}
\end{center}
  \caption{Additional comparison of expanded graphs generated by our model (GEMS) v/s baseline SceneGraphGen on Visual Genome seed graphs. 
  }
\label{fig:baselines5}
\end{figure*}

\begin{figure*}
\begin{center}
\includegraphics[width=\linewidth]{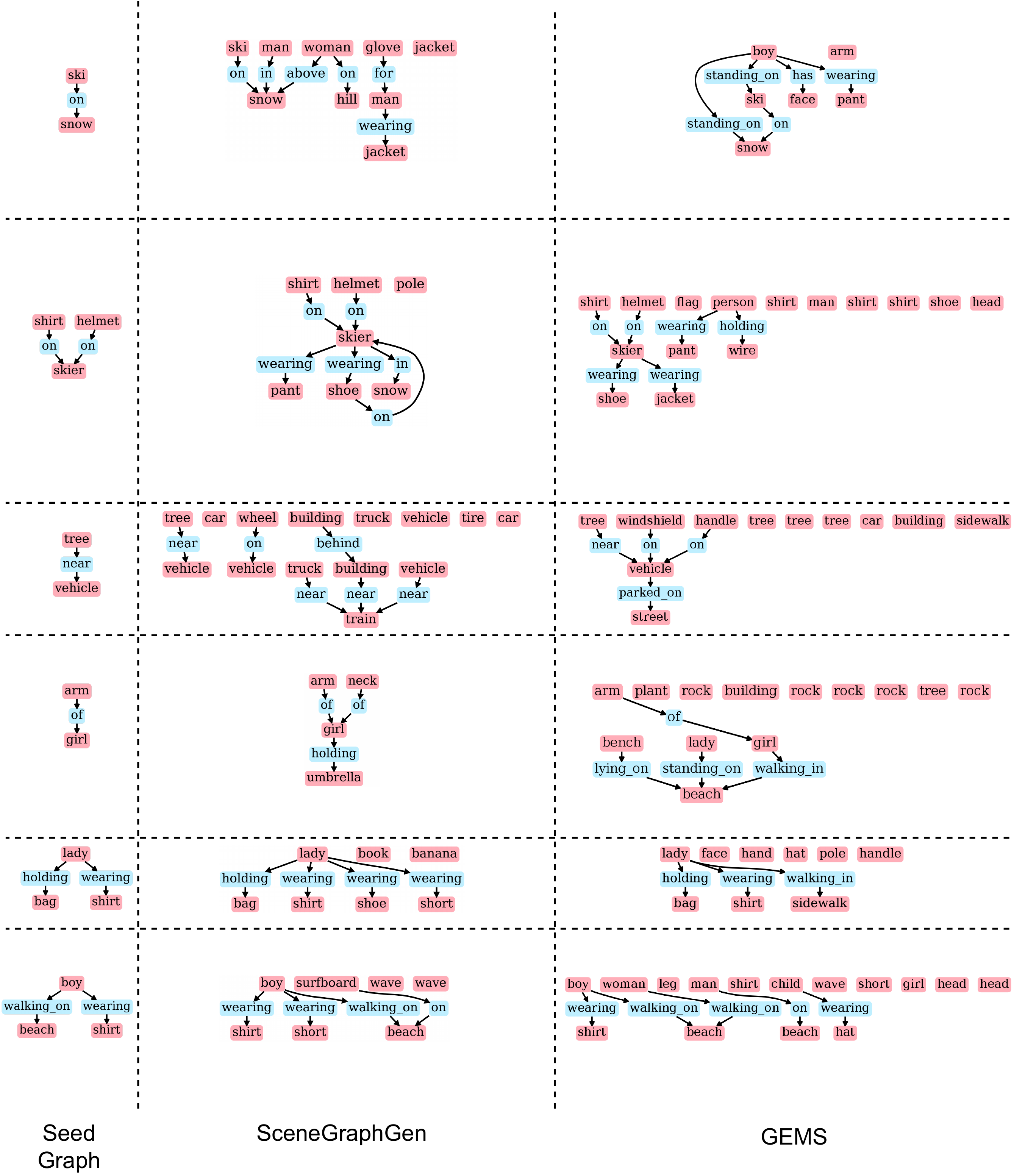}
\end{center}
  \caption{Additional comparison of expanded graphs generated by our model (GEMS) v/s baseline SceneGraphGen on Visual Genome seed graphs. 
  }
\label{fig:baselines6}
\end{figure*}

In figures \ref{fig:baselines1}, \ref{fig:baselines2} and \ref{fig:baselines5}, \ref{fig:baselines6} we show additional results of graphs expanded by our model (GEMS) on Visual Genome seed graphs and compare them with baselines GraphRNN* (GraphRNN with $max\_prev\_node=6$) and SceneGraphGen \cite{scenegraphgen2021} respectively.

\begin{figure*}
\begin{center}
\includegraphics[width=0.7\linewidth]{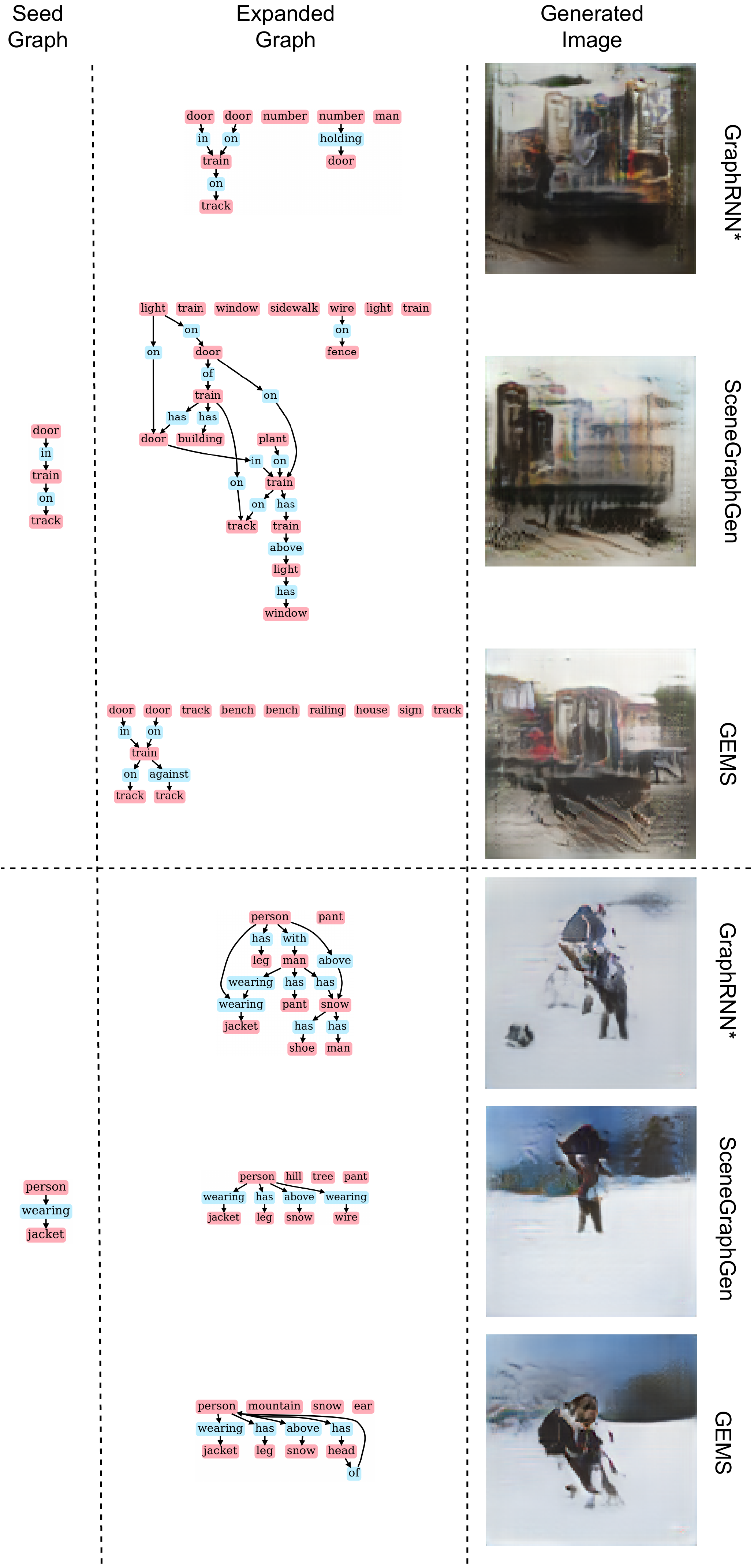}
\end{center}
    \vspace{-0.3cm}
  \caption{Additional comparison of images generated by sg2im \cite{johnson2018image} using expanded scene graphs from seed graphs using our method (GEMS) baselines GraphRNN* and ScenegraphGen. 
  }
\label{fig:baselines3}
\end{figure*}

\begin{figure*}
\begin{center}
\includegraphics[width=0.7\linewidth]{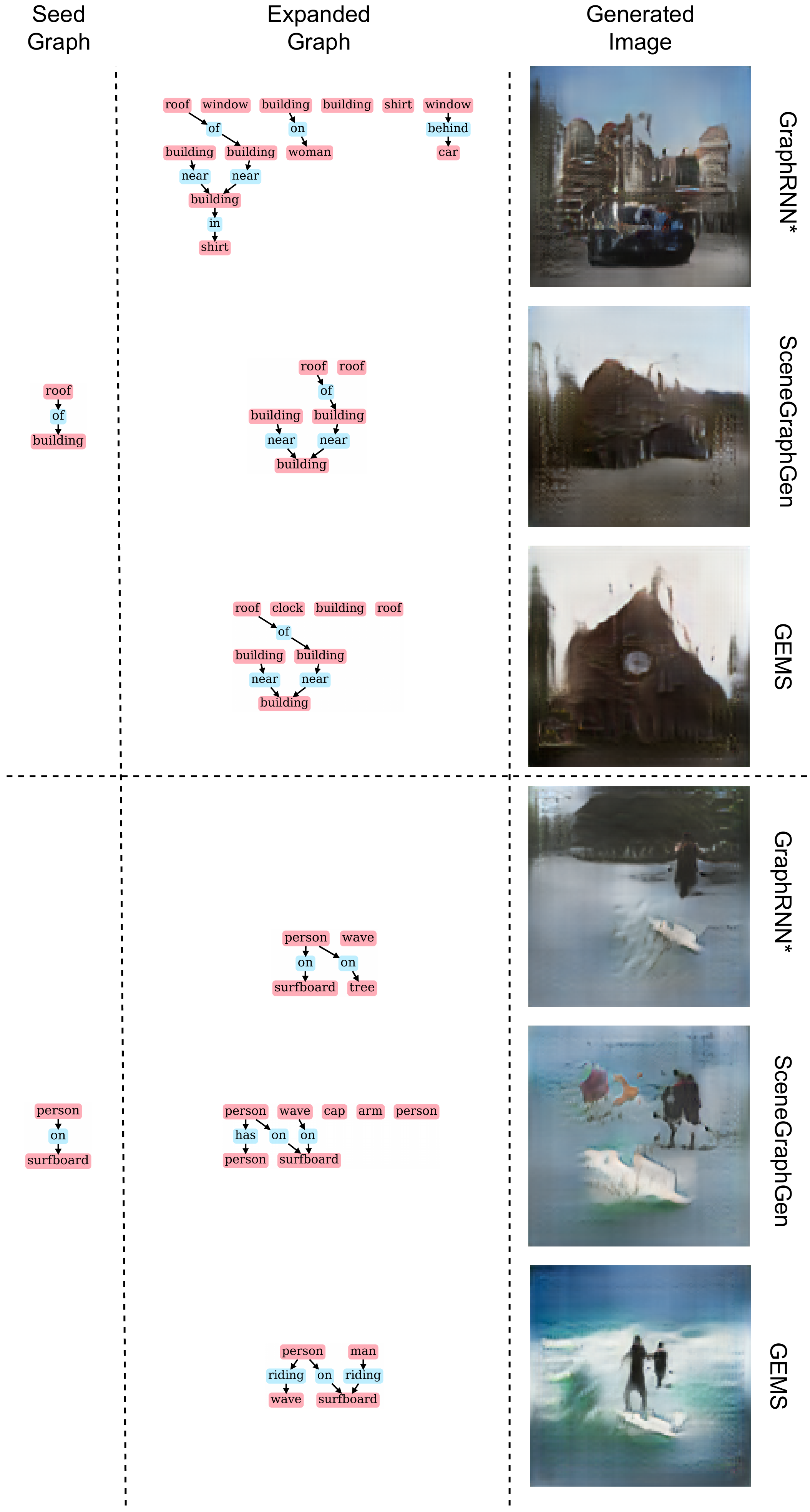}
\end{center}
  \caption{Additional comparison of images generated by sg2im \cite{johnson2018image} using expanded scene graphs from seed graphs using our method (GEMS) baselines GraphRNN* and ScenegraphGen. 
  }
\label{fig:baselines4}
\end{figure*}

In figure \ref{fig:baselines3}, \ref{fig:baselines4} we show additional results of graphs expanded by our model (GEMS) on Visual Genome seed graphs and the corresponding 64x64 images generated by sg2im \cite{johnson2018image} (with the pretrained model for Visual Genome dataset) using these expanded scene graphs against baselines of GraphRNN* (GraphRNN with $max\_prev\_node=6$) and SceneGraphGen \cite{scenegraphgen2021}.

\end{document}